\documentclass{article}

\usepackage[preprint, nonatbib]{neurips_2024}

\usepackage[utf8]{inputenc} 
\usepackage[T1]{fontenc}    
\usepackage{hyperref}       
\usepackage{url}            
\usepackage{booktabs}       
\usepackage{amsfonts}       
\usepackage{nicefrac}       
\usepackage{microtype}      
\usepackage[table]{xcolor}         
\usepackage{subcaption}
\usepackage{graphicx}

\usepackage{tikz}
\usepackage{pgfplots}
\usepackage{amsmath} 
\def\fig#1{Fig. \ref{#1}}
\def\h#1#2{h_{#1}^{#2}}
\makeatletter
\def\Wij{\@ifnextchar\bgroup{\Wij@witharg}{\Wij@noarg}}
\def\dWij{\@ifnextchar\bgroup{\dWij@witharg}{\dWij@noarg}}
\def\DWij{\@ifnextchar\bgroup{\DWij@witharg}{\DWij@noarg}}
\def\Wij@witharg#1{W_{ij}^{(#1)}}
\def\dWij@witharg#1{\delta W_{ij}^{(#1)}}
\def\DWij@witharg#1{\Delta W_{ij}^{(#1)}}
\def\Wij@noarg{W_{ij}}
\def\dWij@noarg{\delta W_{ij}}
\def\DWij@noarg{\Delta W_{ij}}
\makeatother
\def\ltwo#1{\lVert #1 \rVert_2}

\title{SNAP: Stopping Catastrophic Forgetting in Hebbian Learning with Sigmoidal Neuronal Adaptive Plasticity}

\author{%
  Tianyi Xu\thanks{Equal contribution.}, Patrick Zheng\footnotemark[1], Shiyan Liu\thanks{Code is available at: \url{https://github.com/lshiyan/biological-deep-learning/tree/fundamentals_sgd}},  Sicheng Lyu \\
  McGill University\\
  \texttt{\{tianyi.xu2, patrick.zheng, shiyan.liu2, sicheng.lyu\}@mail.mcgill.ca} \\
  \And
  Isabeau Prémont-Schwarz\thanks{Corresponding author} \\
  McGill University\\
  \texttt{isabeau.premont-schwarz@mcgill.ca} \\
}

\begin{document}
\maketitle
\begin{abstract}
Artificial Neural Networks (ANNs) suffer from catastrophic forgetting, where the learning of new tasks causes the catastrophic forgetting of old tasks. Existing Machine Learning (ML) algorithms, including those using Stochastic Gradient Descent (SGD) and Hebbian Learning typically update their weights linearly with experience i.e., independently of their current strength. This contrasts with biological neurons, which at intermediate strengths are very plastic, but consolidate with Long-Term Potentiation (LTP) once they reach a certain strength. We hypothesize this mechanism might help mitigate catastrophic forgetting. We introduce Sigmoidal Neuronal Adaptive Plasticity (SNAP) an artificial approximation to Long-Term Potentiation for ANNs by having the weights follow a sigmoidal growth behaviour allowing the weights to consolidate and stabilize when they reach sufficiently large or small values. We then compare SNAP to linear weight growth and exponential weight growth and see that SNAP completely prevents the forgetting of previous tasks for Hebbian Learning but not for SGD-base learning. 
\end{abstract}

\section{Introduction}

Continual learning is a remarkable human ability that allows for the sequential acquisition of new tasks while minimizing disruptions to previously learned knowledge. This capability supports the accumulation of skills and information throughout one’s lifetime. However, artificial neural networks, particularly those trained with standard gradient descent methods, struggle with this process, often suffering from catastrophic forgetting—where new learning significantly degrades performance on earlier tasks \cite{parisi2019continual}. 

Traditional deep learning approaches assume training data are independent and identically distributed (i.i.d.), which clashes with the sequential nature of continual learning \cite{french1999catastrophic}. Even in humans, training on i.i.d. data can lead to suboptimal performance, suggesting that conventional models may not effectively replicate the learning dynamics observed in biological systems \cite{carvalho2014putting, flesch2018comparing}.

Both Hebbian Learning and Stochastic Gradient Descent (SGD) trained models employ what we will call linear weight growth. By linear weight growth we mean that the size of the of the weight update is independent of the size of the weights or the amount of weight updates already done\footnote{Having a learning rate which decreases over time (learning rate decay) does take into account the amount of experience seen, but as the learning rate applies to all neurons and synapses equally, it simply amounts to a global consolidation of learning and a freezing of all learning.} (cf. \fig{fig:linear-weight-growth}). In contrast, the brain’s learning process includes distinct phases: initial rapid learning, where new information is quickly acquired but also easily forgotten, and Long-Term Potentiation (LTP), where synaptic connections stabilize. 
During LTP, learning does not strengthen further, but the acquired knowledge becomes resistant to forgetting. showcasing a non-linear pattern of neuroplasticity.

Inspired by these biological insights, we propose Sigmoidal Neuronal Adaptive Plasticity (SNAP) a novel approach to weight growth, to approximating the brain’s learning and consolidation phases by have the weights grow like a sigmoid. This method captures the essence of LTP, where synaptic weights initially grow rapidly but eventually stabilize, maintaining learned information without further increases in strength. We explore two variants of SNAP. In the first, synapse-wise SNAP, or s-SNAP, plasticity is defined at the level of each individual weight and depends only on the value of the weight. Thus, in s-SNAP, a neuron can have weights with different levels of plasticity. In the second, neuron-wise SNAP, or n-SNAP, plasticity is defined at the level of the neuron and depends on the all the input weights. Thus in n-SNAP all the input weights of a neuron have the same plasticity, but different neurons have varying levels of plasticity. 

We test out different weight growth behaviour: linear (what is normally done), sigmoidal (to imitate LTP), and exponential. We test both neuron-wise and synapse-wise plasticity, and both Hebbian and SGD based learning. We find that in the i.i.d. condition all types of weight growth behaviour can achieve good performance. In the sequential learning condition however, sigmoidal weight growth prevents catastrophic forgetting in Hebbian Learning but not in SGD based learning where it only slightly reduces catastrophic forgetting. To the best of our knowledge, this is the first time that catastrophic forgetting has been solved in Hebbian Learning without making use of replay (or pseudo-replay) mechanisms. 

\section{Related Works}
\subsection{Catastrophic Forgetting}
The challenge of continual learning has been extensively studied, with various approaches proposed for networks with fixed capacity. These methods can generally be classified into three main categories: replay, regularization-based, and parameter isolation \cite{catastrophic-survey}. Replay methods, inspired by the brain’s episodic replay during sleep and rest, periodically revisit stored samples during or after learning a new task, effectively rehearsing previous knowledge \cite{replay-reasoning}. Previous attempts at avoiding catastrophic forgetting with Hebbian Learning have used pseudo-replay methods \cite{HebbPseudoReplay}. Regularization-based methods draw inspiration from the brain’s synaptic meta-plasticity, adjusting each synaptic weight based on its estimated importance, as determined by other techniques \cite{regularization-reasoning}. Lastly, parameter isolation methods allocate specific model parameters to different tasks, allowing for specialization without interference \cite{parameter-isolation-reasoning}. SNAP effectively uses the weight strength to determine importance and can be considered an implicit regularization-based method. To the best of our knowledge, it is the first method which does not require any additional machinery which explicitly keeps track of previous learning, like stored past data, pseudo-patterns, or past weights to prevent catastrophic forgetting.

\section{Theory}

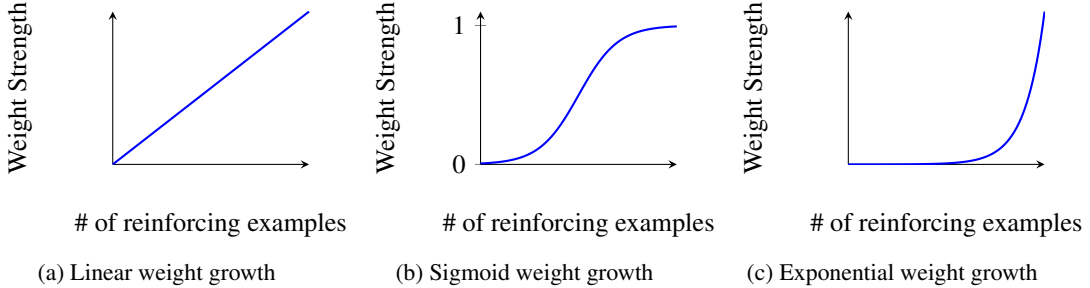
\begin{figure}[!htbp]
    \centering
    \begin{subfigure}[b]{0.30\textwidth} 
        \centering
        \begin{tikzpicture}
            \begin{axis}[
                width=\textwidth, 
                xlabel={\# of reinforcing examples},
                ylabel={Weight Strength},
                xtick=\empty,
                ytick=\empty,
                axis lines=left,
                domain=0:10,
                samples=100
            ]
            \addplot[thick, blue] {x};
            \end{axis}
        \end{tikzpicture}
        \caption{Linear weight growth}
        \label{fig:linear-weight-growth}
    \end{subfigure}
    \hfill
    \begin{subfigure}[b]{0.30\textwidth}
        \centering
        \begin{tikzpicture}
            \begin{axis}[
                width=\textwidth, 
                xlabel={\# of reinforcing examples},
                ylabel={Weight Strength},
                xtick=\empty,
                ytick={0, 1},
                yticklabels={0, 1},
                axis lines=left,
                domain=0:10,
                samples=100,
                ymin=0, ymax=1.1,
                xmin=0
            ]
            \addplot[thick, blue] {1 / (1 + exp(-x + 5))};
            \end{axis}
        \end{tikzpicture}
        \caption{Sigmoid weight growth}
        \label{fig:Sigmoid-weight-growth}
    \end{subfigure}
    \hfill
    \begin{subfigure}[b]{0.30\textwidth}
        \centering
        \begin{tikzpicture}
            \begin{axis}[
                width=\textwidth, 
                xlabel={\# of reinforcing examples},
                ylabel={Weight Strength},
                xtick=\empty,
                ytick=\empty,
                axis lines=left,
                domain=0:10,
                samples=100
            ]
            \addplot[thick, blue] {exp(x)};
            \end{axis}
        \end{tikzpicture}
        \caption{Exponential weight growth}
        \label{fig:Exponential-weight-growth}
    \end{subfigure}
    \caption{Illustrations of different weight growth behaviours}
    \label{fig:weight-growth}
\end{figure}

Let $W_{ij}$ be the synaptic weight linking presynaptic neuron $j$ to postsynaptic neuron $i$ in an ANN. For example, in the case of an MLP (without bias terms) with ReLU activation where neuron $i$ is on layer $l+1$ with activation value $h_i^{l+1}$ and neuron $j$, on layer $l$, has activation value $h_j^{l}$ we have $\h{i}{l+1} = \text{ReLU}(\sum_j W_{ij} \h{j}{l})$. Let us also assume for the sake of simplicity that the weights are positive. Given that the weight is initialized with value $\Wij{0}$ and given a learning algorithm and data which give a weight update $\dWij{t}$ at training step $t$, then with standard machine learning we have that the weight at time $T$ is 
\begin{align}
    \Wij{T} = \Wij{0} + \sum_{t=1}^T \dWij{t}.
    \label{eq:linear_growth_integral}
\end{align}
This is what we call linear weight growth because the weight grows linearly in the $\dWij{t}$'s. This is represented diagrammatically in \fig{fig:linear-weight-growth}.

In order to have LTP-inspired adaptive plasticity, we wish to modify the synaptic strength growth to have the following properties:
\begin{enumerate}
    \item Limited Growth: Synaptic strength does not grow to infinity, but plateaus after reaching a large enough value.
    \item Consolidation: Once the synaptic strength reaches it's plateau, it is difficult for further training to affect it's value.
\end{enumerate}

There are potentially many ways to achieve the above two properties, but having a weight growth behaviour which plateaus after reaching a certain value is the path we wish to explore in this paper. A plateau prevents the weight from growing to infinity, but also, since the slope tends towards zero, it means that further training will barely have an effect on the synaptic weight value. To implement this we choose the sigmoid function (\fig{fig:Sigmoid-weight-growth}).  This means that we wish to change \eqref{eq:linear_growth_integral} to 
\begin{align}
    \Wij{T} = \sigma\left(\Wij{0} + \sum_{t=1}^T \dWij{t}\right), 
    \label{eq:sigmoid_growth_integral}
\end{align}
where $\sigma(x) = \frac{1}{1+e^{-x}}$ is the sigmoid function, as represented diagrammatically in \fig{fig:Sigmoid-weight-growth}. From \eqref{eq:sigmoid_growth_integral} we have that 
\begin{align}
    \Wij{T} - \Wij{T-1}\approx \left(1 - \Wij{T-1}\right)\Wij{T-1}\dWij{T}. 
    \label{eq:sigmoid_growth_der_diff}
\end{align}

Similarly, if instead of wanting sigmoidal growth we had wanted exponential weight growth, instead of \eqref{eq:sigmoid_growth_der_diff} we would have gotten
\begin{align}
    \Wij{T} - \Wij{T-1}\approx \Wij{T-1}\dWij{T}. 
    \label{eq:exp_growth_der_diff}
\end{align}
For more details on the derivation of \eqref{eq:sigmoid_growth_der_diff} and \eqref{eq:exp_growth_der_diff} see \ref{app:WeightGrowthDerivation}. 

Accommodating negative weights by using absolute values in \eqref{eq:sigmoid_growth_der_diff} we can turn any learning rule which provides weight changes $\dWij$ into a rule which has sigmoidal weight growth by applying weight changes $\DWij$ instead where
\begin{align}
    \DWij = |\Wij|(1 - |\Wij|) \dWij. 
    \label{eq:s-snap}
\end{align}

In the above, the plasticity of each synapse can be different and is given by $|\Wij|(1 - |\Wij|)$. Modifying the weight updates as in \eqref{eq:s-snap} gives us synapse-wise SNAP.

But it may be useful to have the plasticity of all incoming synapses to a neuron be tied. We can do this simply by asking that instead of $\Wij$ needing to follow a sigmoidal growth pattern,  $\ltwo{W_i} = \sqrt{\sum_j\Wij^2}$ needs to follow a sigmoidal growth pattern. And this can be achieved simply by modifying the weight change to
\begin{align}
    \DWij = \ltwo{W_i}(1 - \ltwo{W_i}) \dWij. 
    \label{eq:n-snap}
\end{align}

In the above, the plasticity of each input synapse at neuron $i$ is the same and is given by $\ltwo{W_i}(1 - \ltwo{W_i})$. Modifying the weight updates as in \eqref{eq:n-snap} gives us neuron-wise SNAP.

\begin{table}[h]
\centering
\begin{tabular}{l|ccc}
\toprule
\rowcolor[gray]{0.8} & \textbf{Linear} & \textbf{Sigmoidal} & \textbf{Exponential} \\
\midrule
\rowcolor[gray]{0.9} \textbf{Synapse-wise} & $\DWij = \dWij$ & $\DWij = |\Wij|(1 - |\Wij|) \dWij$ & $\DWij = |\Wij|\dWij$ \\
\rowcolor[gray]{0.9} \textbf{Neuron-wise} & $\DWij = \dWij$ & $\DWij = \ltwo{W_i}(1 - \ltwo{W_i}) \dWij$ & $\DWij =\ltwo{W_i} \dWij$ \\
\bottomrule
\end{tabular}
\caption{Given a learning rule which given weight updates $\delta W_{ij}$, applying the modified weight updates $\Delta W_{ij}$ instead, will allow for linear, sigmoidal, or exponential growth in either the weight values (synapse-wise) or in the input-weight norms (neuron-wise).}
\label{tab:growth-eqns}
\end{table}

\section{Experiments}

\subsection{Experiment 1: I.I.D. Data}
We train an MLP with one hidden layer (cf. section \ref{app:architecture} for more details on the architecture) using either Hebbian Learning (cf. section \ref{app:HL} for more details on our implementation of Hebbian Learning) or SGD with cross-entropy loss, on i.i.d. datasets (MNIST and FashionMNIST).

\begin{figure}[htbp]
  \centering
  \begin{subfigure}[b]{0.48\textwidth}
    \centering
    \includegraphics[width=\textwidth]{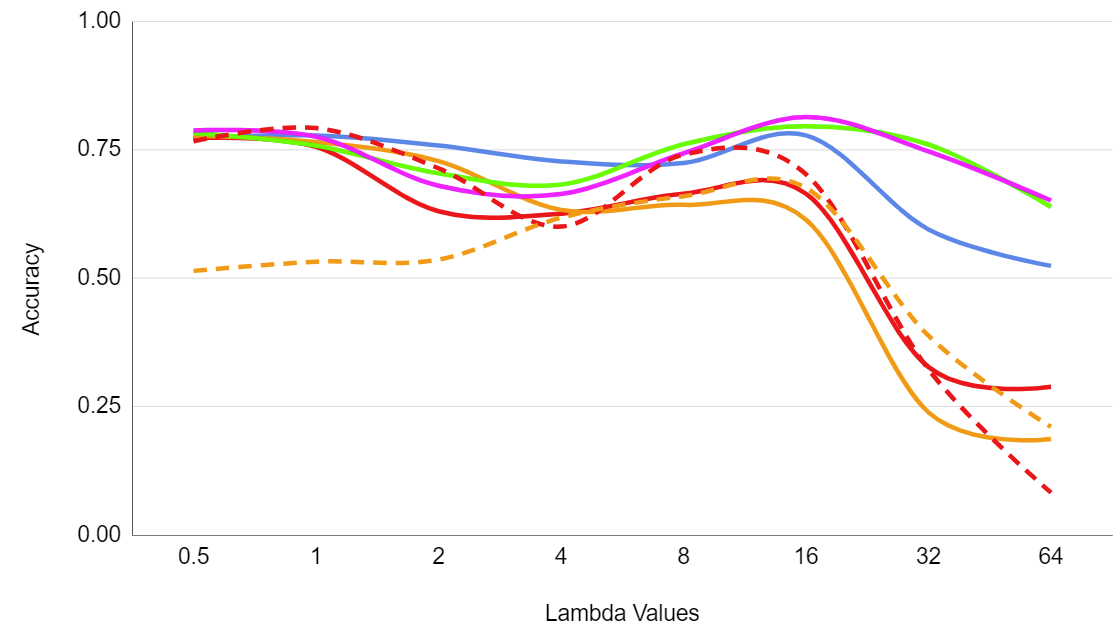}
    \caption{MNIST}
    \label{fig:Baseline_MNIST_Model_Performance}
  \end{subfigure}
  \hfill
  \begin{subfigure}[b]{0.48\textwidth}
    \centering
    \includegraphics[width=\textwidth]{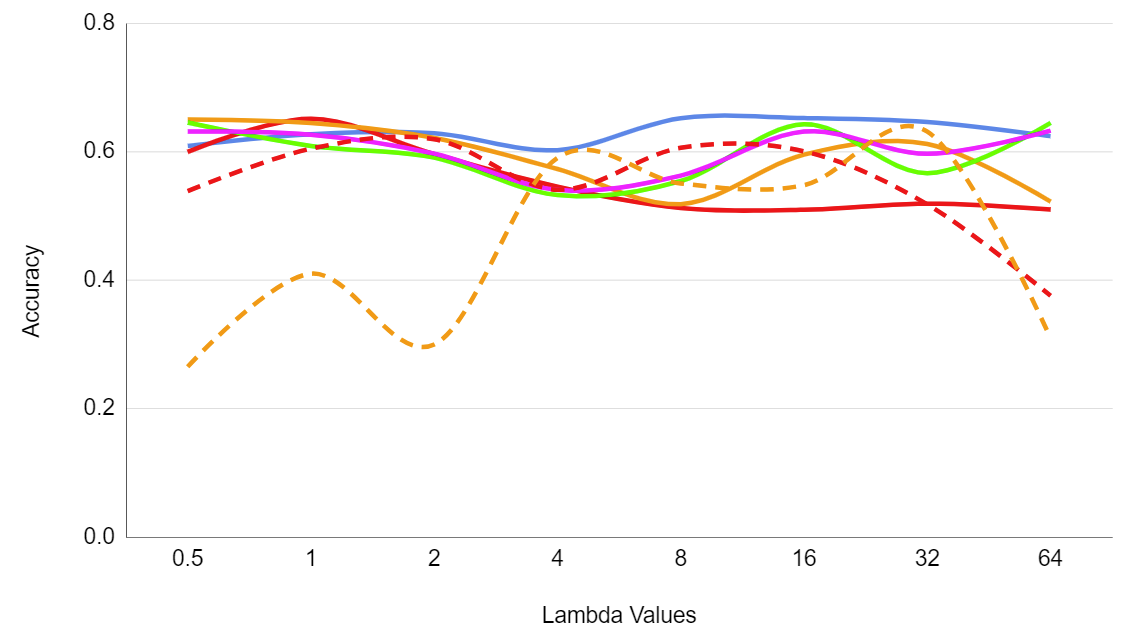}
    \caption{FashionMNIST}
    \label{fig:Baseline_FMNIST_Model_Performance}
  \end{subfigure}
  
  \vspace{1em} 
  
  \begin{subfigure}[b]{1.0\textwidth}
    \centering
    \includegraphics[width=\textwidth]{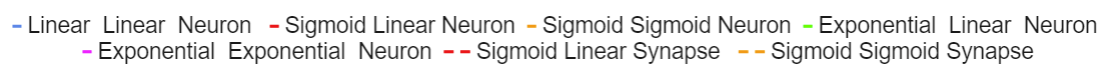}
    \caption{Legend}
    \label{fig:Legend}
  \end{subfigure}
  
  \caption{Accuracy of Hebbian MLP on \textbf{i.i.d. datasets} MNIST and FashionMNIST as a function of lateral inhibition hyperparameter $\lambda$ and the type of weight growth. \textbf{Legend:} the first word is the weight growth type of the hidden layer and the second word is the weight growth type of the output layer. So that for example, red lines represent sigmoidal weight growth for the hidden layer and linear weight growth for the output layer. The solid lines denote neuron-wise weight growth while the dotted lines denote synapse-wise weight growth (cf. table \ref{tab:growth-eqns}).}
  \label{fig:Baseline_Model_Performance}
\end{figure}

For Hebbian Learning, we can see from \fig{fig:Baseline_Model_Performance} that as long as we choose the right hyperparameter $\lambda$ (which controls the strength of lateral inhibition cf. section \ref{app:HL} for more details) all types of weight growth can performe roughly equally well in the i.i.d. setting. The same is true for the SGD trained network (cf. table \ref{tab:sgd-baseline}).

\subsection{Experiment 2: Sequential Task Learning}
\begin{figure}[htbp]
  \centering
  \begin{subfigure}[b]{0.48\textwidth}
    \centering
    \includegraphics[width=\textwidth]{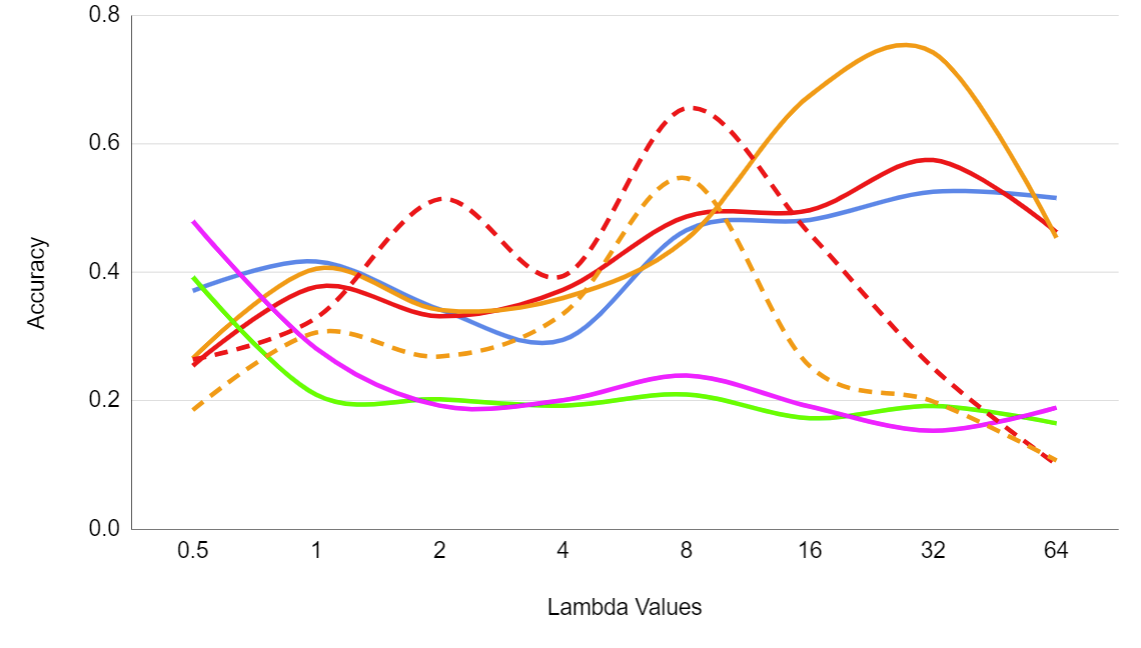}
    \caption{MNIST}
    \label{fig:Forget_MNIST_Model_Performance}
  \end{subfigure}
  \hfill
  \begin{subfigure}[b]{0.48\textwidth}
    \centering
    \includegraphics[width=\textwidth]{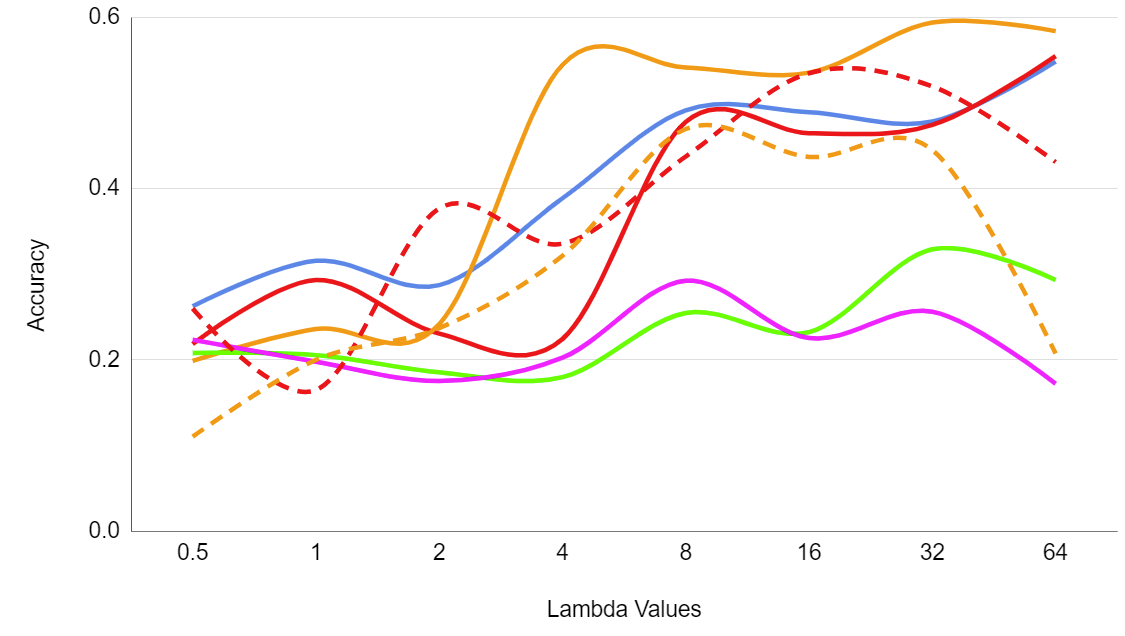}
    \caption{FashionMNIST}
    \label{fig:Forget_Fashion_MNIST_Model_Performance}
  \end{subfigure}
  \vspace{1em} 
  
  \begin{subfigure}[b]{1.0\textwidth}
    \centering
    \includegraphics[width=\textwidth]{ExperimentGraphs/legend_function_type.png}
    \caption{Legend}
    \label{fig:Legend}
  \end{subfigure}
  \caption{Accuracy of Hebbian MLP on the \textbf{sequential task} versions of MNIST and FashionMNIST as a function of lateral inhibition hyperparameter $\lambda$ and the type of weight growth. \textbf{Legend:} the first word is the weight growth type of the hidden layer and the second word is the weight growth type of the output layer. So that for example, red lines represent sigmoidal weight growth for the hidden layer and linear weight growth for the output layer. The solid lines denote neuron-wise weight growth while the dotted lines denote synapse-wise weight growth (cf. table \ref{tab:growth-eqns}).}
  \label{fig:Forget_Model_Performance}
\end{figure}

To test sequential task learning we turn both MNIST and FashionMNIST into sequential tasks by sequentially training the model on five tasks where task 1 trains only on images of classes 0 and 1, task 2 trains only on images of classes 2 and 3, and so forth until task 5 which trains only on images of classes 8 and 9. The model switches to the next task when it reaches 80\% accuracy\footnote{We chose the value of 80\% accuracy because with the network sizes with which we experimented, 80\% is close to, but slightly less than the top accuracies which we achieve.} or 35 epochs, whichever comes first. \fig{fig:Forget_Model_Performance} shows the average test accuracy across all classes (0-9) after training on all tasks sequentially for Hebbian models with different lateral inhibition strengths. 

Models with sigmoidal weight growth in their hidden layers achieve the highest average test accuracies by reducing catastrophic forgetting through a consolidation phase, as shown in Figure \ref{fig:Forget_Model_Performance}. Exponential growth models, despite excelling in I.I.D. experiments, perform the worst due to rapid learning and forgetting, while linear growth models show average performance with moderate forgetting. 

\begin{figure}[htbp]
  \centering
  \begin{subfigure}[t]{0.32\textwidth}
    \centering
    \includegraphics[width=\textwidth]{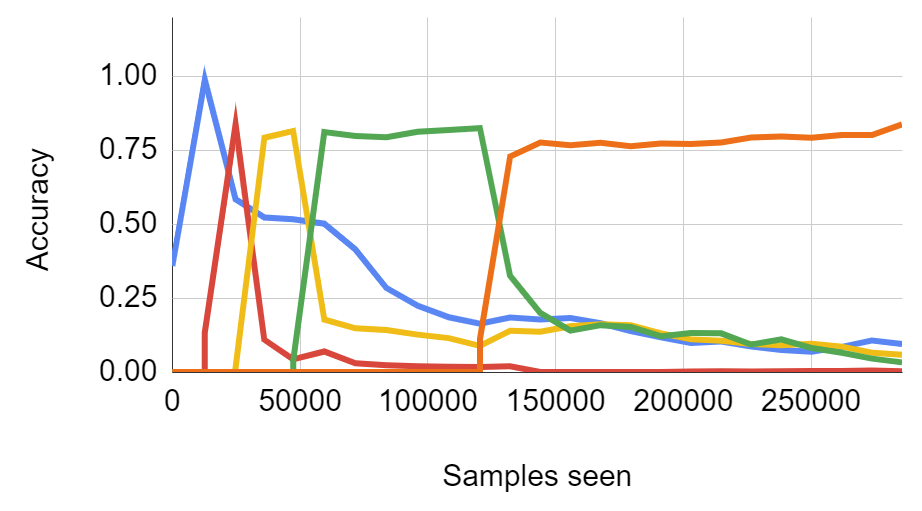}
    \caption{Exponential Hidden and Classification Layer, Neuron-wise, $\lambda = 8$}
    \label{fig:MNIST-FORGET-EXP-EXP-NEURON-8}
  \end{subfigure}
  \hfill
  \begin{subfigure}[t]{0.32\textwidth}
    \centering
    \includegraphics[width=\textwidth]{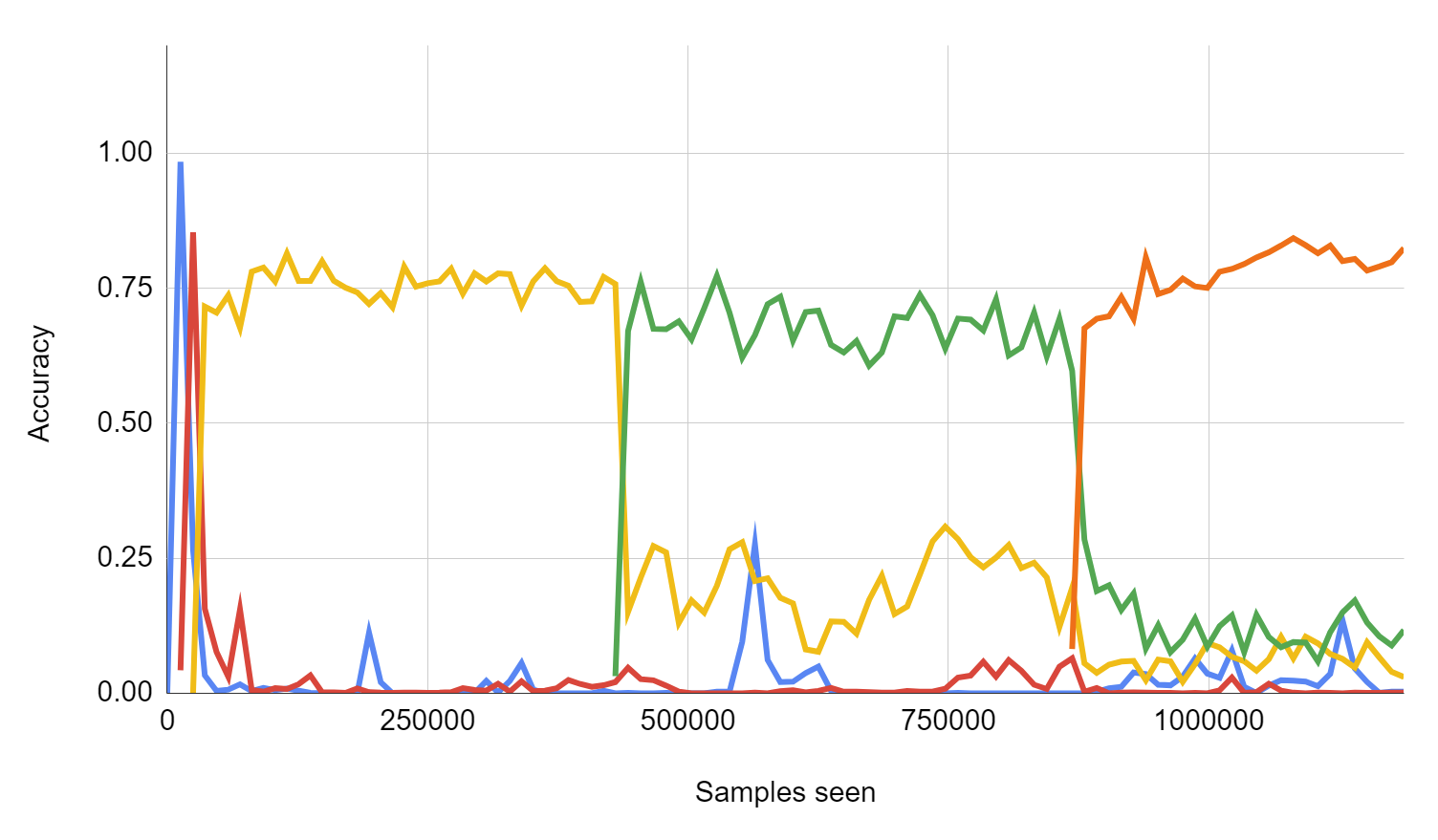}
    \caption{Exponential Hidden, Linear Classification Layer, Neuron-wise, $\lambda = 16$}
    \label{fig:MNIST-FORGET_EXP_LINEAR_16}
  \end{subfigure}
    \hfill
  \begin{subfigure}[t]{0.32\textwidth}
    \centering
    \includegraphics[width=\textwidth]{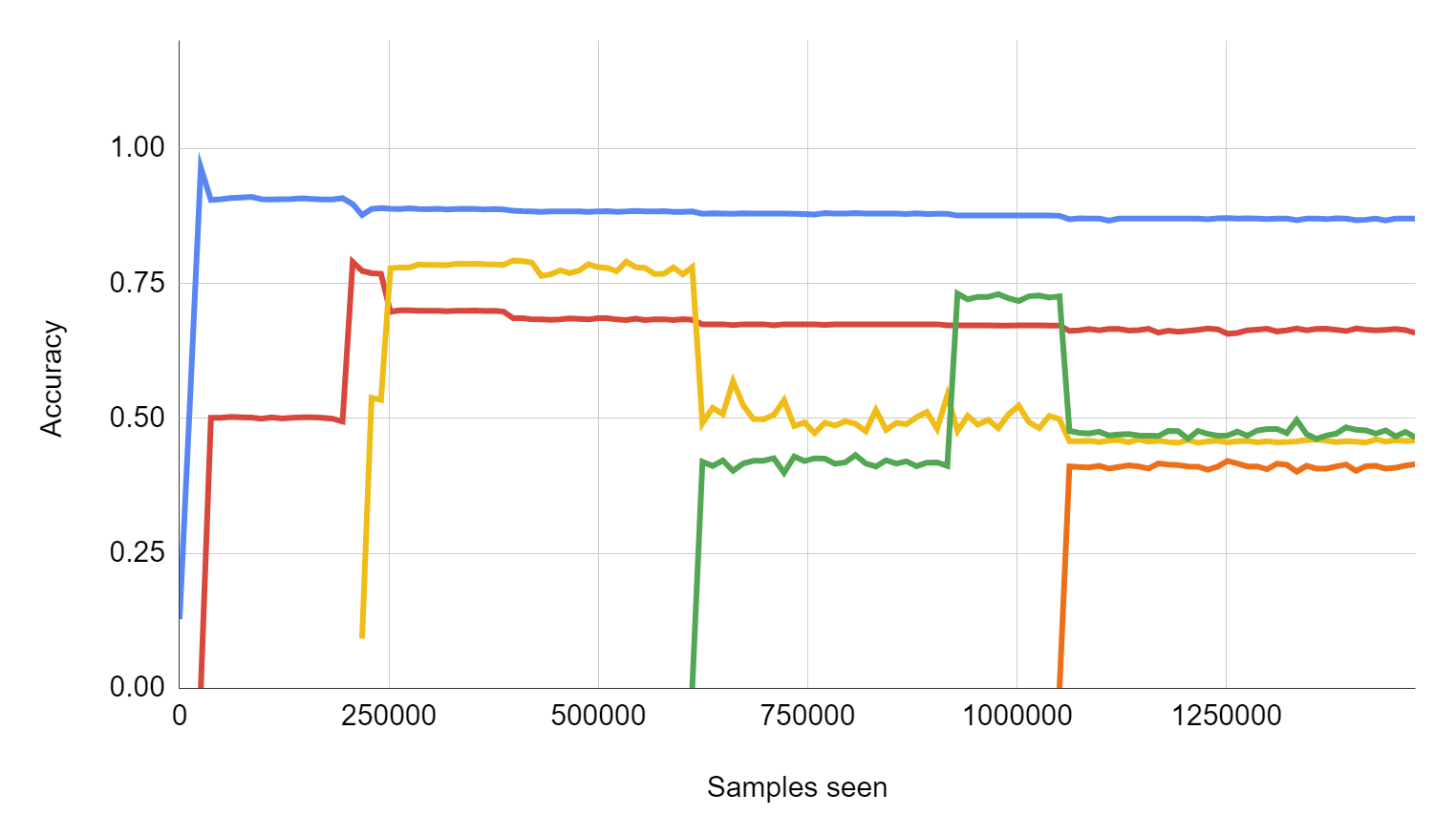}
    \caption{Sigmoid Hidden and Classification Layer, Neuron-wise, $\lambda = 32$}
    \label{fig:MNIST-FORGET_SIG_LINEAR_NEURON_32}
  \end{subfigure}
  \vfill
  \begin{center}
  \begin{subfigure}[t]{0.32\textwidth}
    \centering
    \includegraphics[width=\textwidth]{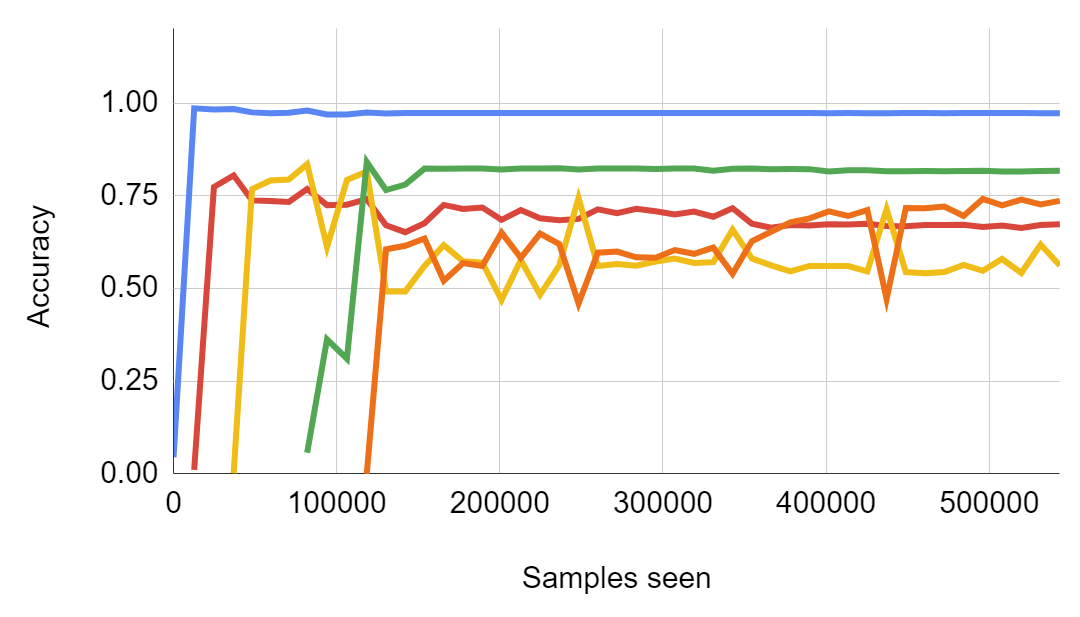}
    \caption{Sigmoid Hidden and Classification Layer, Neuron-wise, $\lambda = 32$}
    \label{fig:MNIST-FORGET_SIG_SIG_NEURON_32}
  \end{subfigure}
  \hspace{0.05\textwidth}
  \begin{subfigure}[t]{0.32\textwidth}
    \centering
    \includegraphics[width=\textwidth]{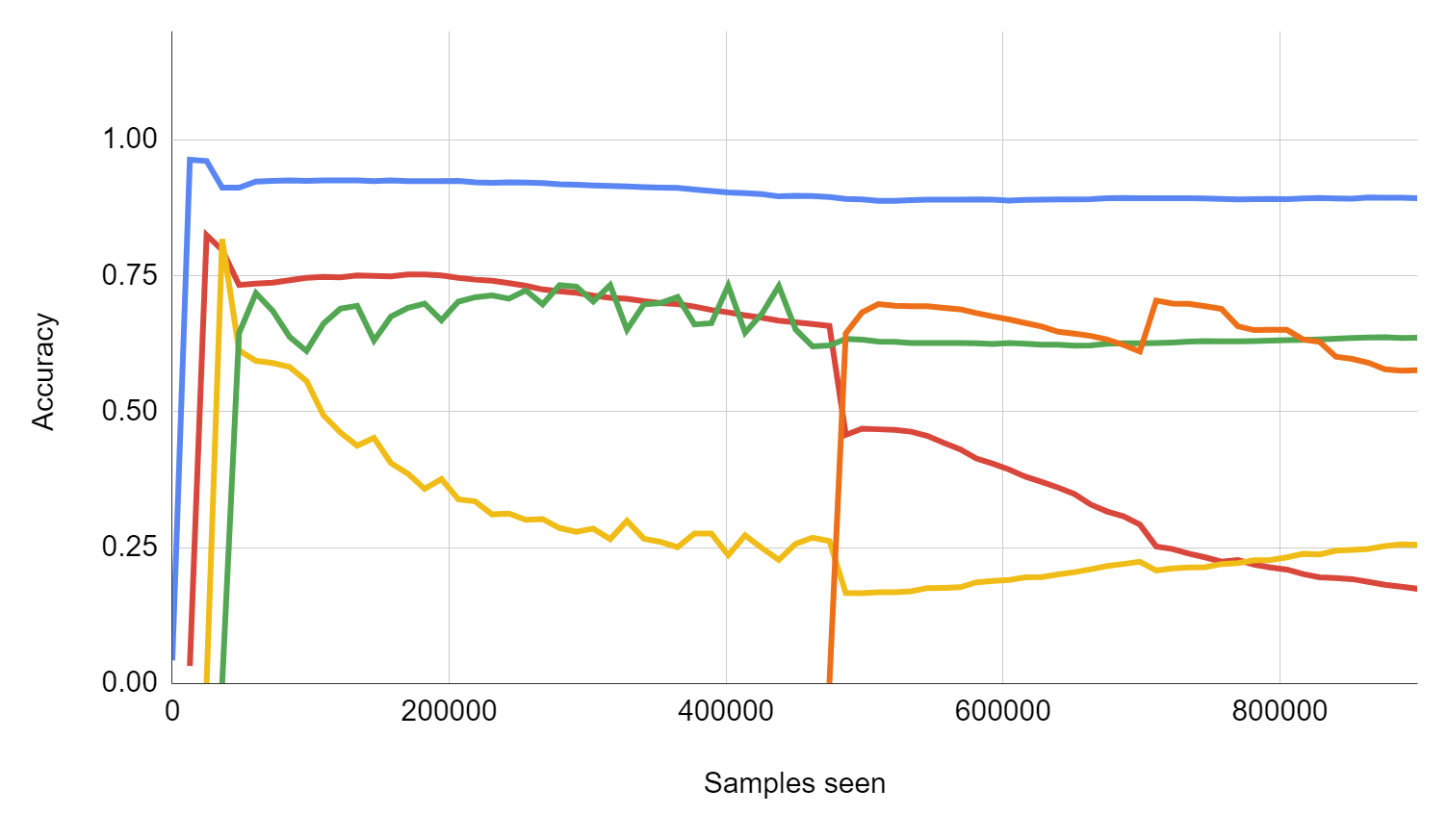}
    \caption{Linear Hidden and Classification Layer, $\lambda = 32$}
    \label{fig:MNIST-FORGET_LINEAR_LINEAR_NEURON_32}
  \end{subfigure}
  \end{center}
  \begin{subfigure}[b]{1.0\textwidth}
    \centering
    \includegraphics[width=\textwidth]{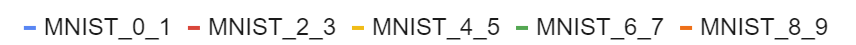}
    \caption{Legend}
    \label{fig:Legend}
  \end{subfigure}
  \caption{Comparison of top-performing Hebbian models on sequential MNIST task learning experiment with various hidden and classification layer weight growth types. \textbf{Legend:} Each line represents a specific class's test accuracy. For instance, the blue line represents the test accuracy of MNIST classes 0 and 1 throughout the experiment.}
  \label{fig:mnist_hebb_combined_Forget_Model_Performance}
\end{figure}

Figures \ref{fig:MNIST-FORGET-EXP-EXP-NEURON-8}, \ref{fig:MNIST-FORGET_SIG_LINEAR_NEURON_32}, \ref{fig:MNIST-FORGET_SIG_SIG_NEURON_32}, \ref{fig:MNIST-FORGET_LINEAR_LINEAR_NEURON_32}, and \ref{fig:MNIST-FORGET_EXP_LINEAR_16} present the best-performing Hebbian models on sequential MNIST learning (cf. \fig{fig:fmnist_hebb_top_performance} for the same results on FashionMNIST) from each growth type: exponential-exponential, exponential-linear, sigmoid-linear, sigmoid-sigmoid, and linear-linear, respectively, where sigmoid-linear means that the hidden layer has sigmoidal weight growth while the outpul layer has linear weight growth.  

Importantly, we can see from Fig. \ref{fig:MNIST-FORGET_LINEAR_LINEAR_NEURON_32}  and \ref{fig:fmnist_hebb_FORGET_LINEAR_LINEAR_NEURON_16} that even the best linear growth Hebbian learning also suffers from catastrophic forgetting even if it is not as severe as for SGD trained model (compare with Figs. \ref{fig:FORGET-LINEAR-SGD-MNIST} and \ref{fig:FORGET-LINEAR-SGD-FMNIST}). 

From Figs. \ref{fig:MNIST-FORGET_SIG_SIG_NEURON_32} and \ref{fig:fmnist_hebb_FORGET_SIG_SIG_NEURON_32} we see that for Hebbian learning, neuron-wise Sigmoidal weight growth completely prevents any forgetting of previous tasks while training on new tasks. The same cannot be said of SGD trained networks (cf. Figs \ref{fig:FORGET-SIGMOID-NEURON-SGD-MNIST} and \ref{fig:FORGET-SIGMOID-NEURON-SGD-FMNIST}) even though with sigmoidal growth it forgets slightly more slowly than with linear weight growth. When we look into why it solve catastrophic forgetting for SGD, it is because the network reaches high accuracies before the weights have grown to be close to $1$ in size, i.e. before the weights reach their consolidation phase. 


\section{Conclusions}
In this paper, we investigated sigmoidal weight growth as a neuro-inspired mechanism to prevent catastrophic forgetting in sequential task learning. 

Hebbian learning naturally suffers slightly less than SGD-trained networks from catastrophic forgetting, but still suffers from it enough that it cannot train successfully on many sequential tasks. However, when trained with SNAP, i.e. with sigmoidal weight growth, we have a total protection against the forgetting of previous tasks.

SNAP, while helping slightly with catastrophic forgetting for SGD-trained networks, does not prevent it in this case. We leave it up to future research to see if SNAP can be successfully adapted to SGD.

\section{Acknowledgments}
This research was supported by a grant from \href{https://strongcompute.com/}{Strong Compute} who generously provided us with compute.  

\bibliographystyle{plain}
\bibliography{snap}

\appendix
\section{Supplementary Material}

\subsection{Neural Net Architecture}
\label{app:architecture}
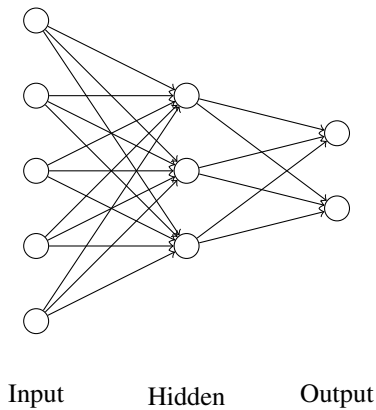
\begin{figure}[h]
    \centering
    \begin{tikzpicture}
        \node[draw, circle] (A) at (0, 2) {};
        \node[draw, circle] (B) at (0, 1) {};
        \node[draw, circle] (C) at (0, 0) {};
        \node[draw, circle] (D) at (0, -1) {};
        \node[draw, circle] (E) at (0, -2) {};

        \node[draw, circle] (F) at (2, 1) {};
        \node[draw, circle] (G) at (2, 0) {};
        \node[draw, circle] (H) at (2, -1) {};

        \node[draw, circle] (I) at (4, 0.5) {};
        \node[draw, circle] (J) at (4, -0.5) {};

        \draw[->] (A) -- (F);
        \draw[->] (A) -- (G);
        \draw[->] (A) -- (H);
        
        \draw[->] (B) -- (F);
        \draw[->] (B) -- (G);
        \draw[->] (B) -- (H);
        
        \draw[->] (C) -- (F);
        \draw[->] (C) -- (G);
        \draw[->] (C) -- (H);
        
        \draw[->] (D) -- (F);
        \draw[->] (D) -- (G);
        \draw[->] (D) -- (H);
        
        \draw[->] (E) -- (F);
        \draw[->] (E) -- (G);
        \draw[->] (E) -- (H);

        \draw[->] (F) -- (I);
        \draw[->] (G) -- (I);
        \draw[->] (H) -- (I);

        \draw[->] (F) -- (J);
        \draw[->] (G) -- (J);
        \draw[->] (H) -- (J);

        \node at (0, -3) {$\text{Input}$};
        \node at (2, -3) {$\text{Hidden}$};
        \node at (4, -3) {$\text{Output}$};
        
    \end{tikzpicture}
    \caption{Simplified Model Configuration}
    \label{fig:simplified-model-config}
\end{figure}

Our models follow a simple architecture, (Figure \ref{fig:simplified-model-config}), that consists of three layers: input, hidden, and output.

\subsection{Notation Introduction}

\begin{table}[h]
    \centering
    \begin{tabular}{|l|l|p{9cm}|}
        \hline
        \textbf{Category} & \textbf{Symbol} & \textbf{Description} \\
        \hline
        Hidden Layer & \(x_i\) & Activity of the presynaptic neuron. \\
        & \(a_i\) & Activity of the postsynaptic neuron. \\
        & \(r_{ij}\) & Computed result term from the learning rule, used for weight updates. \\
        & \(h_i\) & Post lateral inhibition activation of each neuron in the hidden layer. \\
        \hline
        Classification Layer & \(\hat{y}_i\) & Ground truth label for the sample. \\
        & \(\tilde{y}_i\) & Predicted label by the model. \\
        & \(h_i\) & Post lateral inhibition activation from the hidden layer, serving as input. \\
        \hline
        Weight Update & \(\delta W_{ij}\) & Change in weights. \\
        & \(\alpha\) & Learning rate. \\
        & \(f(x)\) & Function used for weight growth (e.g., linear, sigmoid, exponential). \\
        \hline
    \end{tabular}
    \caption{Notation for Hidden Layer, Classification Layer, and Weight Update}
    \label{tab:notation_introduction}
\end{table}

\subsection{Hyperparameter Specification}

\textbf{I.I.D. Experiments Hyperparameters}

For all experiments, our models use the hyperparameters specified in Table \ref{tab:mnist_fmnist_hyperparameters}. We evaluated the optimal values of $\eta$ (as defined in Equation \ref{eq:sanger}) and $\alpha$ for seven weight growth model configurations: linear-linear, sigmoid-linear neuron, sigmoid-linear synapse, exponential-linear neuron, exponential-exponential neuron, sigmoid-sigmoid neuron, and sigmoid-sigmoid synapse. These evaluations were conducted on both MNIST and FashionMNIST datasets.

In these configurations, the first term refers to the growth function applied to the hidden layer, and the second term to the output layer. Growth can be either neuron-wise (applied across all synapses of a neuron) or synapse-wise (applied individually to each synapse). For example, "sigmoid-linear neuron" means sigmoid growth for the hidden layer and linear growth for the output layer, applied on a neuron-wise basis.

For each configuration, we performed the I.I.D. experiments over 10 epochs to identify the best $\eta$ and $\alpha$ pairs based on test accuracy. The identified optimal pairs were then used consistently across all experiments, including both the I.I.D. and Sequential Task experiments, without further re-tuning for the Sequential Task experiments.

The values of $\eta$ tested were 0.7, 0.3, 0.1, 0.03, and 0.001. The values of $\alpha$ tested were 0.3, 0.1, 0.03, 0.01, and 0.003.

The optimal $\eta$ and $\alpha$ pairs for MNIST are detailed in Table \ref{table:mnist-configurations}, while the corresponding values for FashionMNIST are presented in Table \ref{table:fmnist-configurations}.

\begin{table}[h]
    \centering
    \begin{tabular}{|l|c|c|}
        \hline
         \textbf{Hyperparameter} & \textbf{MNIST Experiment} & \textbf{FashionMNIST Experiment} \\
        \hline
        Input Dimension & 784 & 784 \\
        Hebbian Dimension & 64 & 96 \\
        Output Dimension & 10 & 10 \\
        Beta & 0.01 & 0.01 \\
        Initialization & Uniform(0,beta) & Uniform(0,beta) \\
        Batch Size & 1 & 1 \\
        Epochs & 10 & 10 \\
        \hline
    \end{tabular}
    \caption{General Model Hyperparameters for MNIST and FashionMNIST Experiments}
    \label{tab:mnist_fmnist_hyperparameters}
\end{table}

\begin{table}[ht]
\centering
\small 
\setlength{\tabcolsep}{4pt} 
\renewcommand{\arraystretch}{1.2} 

\resizebox{\textwidth}{!}{%
\begin{tabular}{|l|c|c|c|c|c|c|c|c|}
\hline
\textbf{$\lambda$ / Model} & \textbf{L.L.} & \textbf{E.L. Neuron} & \textbf{S.L. Neuron} & \textbf{S.S. Neuron} & \textbf{E.E. Neuron} & \textbf{S.L. Synapse} & \textbf{S.S. Synapse} \\
\hline
\textbf{0.5} 
    & (0.03, 0.01) & (1, 0.001) & (1, 0.01) & (0.3, 0.1) & (1, 0.001) & (1, 0.001) & (0.1, 0.001) \\
\hline
\textbf{1} 
    & (0.1, 0.01) & (1, 0.01) & (0.3, 0.01) & (0.3, 0.03) & (1, 0.001) & (1, 0.001) & (1, 0.001) \\
\hline
\textbf{2} 
    & (0.1, 0.001) & (1, 0.001) & (0.1, 0.01) & (0.3, 0.001) & (1, 0.03) & (1, 0.01) & (0.3, 0.03) \\
\hline
\textbf{4} 
    & (0.03, 0.001) & (1, 0.001) & (0.3, 0.001) & (0.3, 0.001) & (1, 0.001) & (0.7, 0.001) & (0.3, 0.001) \\
\hline
\textbf{8} 
    & (0.01, 0.01) & (1, 0.001) & (0.3, 0.001) & (0.3, 0.001) & (1, 0.001) & (0.1, 0.03) & (0.3, 0.03) \\
\hline
\textbf{16} 
    & (0.01, 0.03) & (1, 0.01) & (0.3, 0.001) & (0.7, 0.001) & (1, 0.01) & (0.3, 0.1) & (0.7, 0.01) \\
\hline
\textbf{32} 
    & (0.01, 0.01) & (0.3, 0.1) & (0.7, 0.1) & (0.7, 0.3) & (0.3, 0.03) & (0.7, 0.1) & (0.7, 0.1) \\
\hline
\textbf{64} 
    & (0.01, 0.01) & (0.1, 0.1) & (0.7, 0.3) & (0.01, 0.1) & (0.1, 0.03) & (0.3, 0.3) & (0.01, 0.001) \\
\hline
\end{tabular}%
}
\caption{Optimal $\eta$ and learning rate pairs for various weight growth model configurations on the FashionMNIST dataset. These values represent the best-performing parameters from the I.I.D. classification experiments and were used consistently across all experiments, including the sequential task learning experiment. Here, L represents linear, S represents sigmoid, and E represents exponential.}
\label{table:fmnist-configurations}
\end{table}

\begin{table}[ht]
\centering
\small 
\setlength{\tabcolsep}{4pt} 
\renewcommand{\arraystretch}{1.2} 

\resizebox{\textwidth}{!}{%
\begin{tabular}{|l|c|c|c|c|c|c|c|c|}
\hline
\textbf{$\lambda$ / Model} & \textbf{L.L.} & \textbf{E.L. Neuron} & \textbf{S.L. Neuron} & \textbf{S.S. Neuron} & \textbf{E.E. Neuron} & \textbf{S.L. Synapse} & \textbf{S.S. Synapse} \\
\hline
\textbf{0.5} 
    & (0.03, 0.01) & (1, 0.001) & (1, 0.01) & (0.3, 0.1) & (1, 0.001) & (1, 0.001) & (0.1, 0.001) \\
\hline
\textbf{1} 
    & (0.1, 0.01) & (1, 0.01) & (0.3, 0.01) & (0.3, 0.03) & (1, 0.001) & (1, 0.001) & (1, 0.001) \\
\hline
\textbf{2} 
    & (0.1, 0.001) & (1, 0.001) & (0.1, 0.01) & (0.3, 0.001) & (1, 0.03) & (1, 0.01) & (0.3, 0.03) \\
\hline
\textbf{4} 
    & (0.03, 0.001) & (1, 0.001) & (0.1, 0.01) & (0.3, 0.001) & (1, 0.001) & (1, 0.03) & (0.1, 0.1) \\
\hline
\textbf{8} 
    & (0.01, 0.01) & (1, 0.001) & (0.1, 0.03) & (0.1, 0.1) & (1, 0.001) & (0.1, 0.001) & (0.1, 0.001) \\
\hline
\textbf{16} 
    & (0.01, 0.03) & (1, 0.01) & (0.003, 0.01) & (0.003, 0.3) & (1, 0.01) & (1, 0.01) & (1, 0.1) \\
\hline
\textbf{32} 
    & (0.01, 0.01) & (0.3, 0.1) & (0.01, 0.01) & (0.01, 0.3) & (0.3, 0.03) & (0.1, 0.3) & (1, 0.1) \\
\hline
\textbf{64} 
    & (0.01, 0.01) & (0.1, 0.1) & (0.001, 0.1) & (0.003, 0.3) & (0.1, 0.03) & (0.03, 0.03) & (0.3, 0.3) \\
\hline
\end{tabular}%
}
\caption{Optimal $\eta$ and learning rate pairs for various weight growth model configurations on the MNIST dataset. These values represent the best-performing parameters from the I.I.D. classification experiments and were used consistently across all experiments, including the sequential task learning experiment. Here, L represents linear, S represents sigmoid, and E represents exponential.}
\label{table:mnist-configurations}
\end{table}

\subsection{Hebbian Learning}
\label{app:HL}
\subsubsection{Lateral Inhibition in Feedforward Propogation}
During forward propagation, lateral inhibition—a mechanism where excited neurons suppress their neighbors—promotes sparse, distinct activations, enhancing contrast in stimuli. Our model uses equation \eqref{eq:lateral_inhibition}, where increasing the \(\lambda\) parameter results in sparser hidden layer neuron activations.
\begin{equation}
a_i = \text{ReLU}\left(\sum_j W_{ij} x_j\right)
\label{eq:activation}
\end{equation}

\begin{equation}
h_i = \left(\frac{a_i}{\max_k(a_k)}\right)^\lambda
\label{eq:lateral_inhibition}
\end{equation}

\subsubsection{Weight Update Mechanism}
Most existing approaches train neural network layers sequentially: first, the initial layer is trained on all the data, and then the outputs from this trained layer are used as inputs to the next layer, and so on. While this method may simplify training, we do not consider it realistic or practical for real-world applications where simultaneous learning is often required. Therefore, in our approach, we train all layers simultaneously, allowing the network to learn in a more integrated and biologically relevant manner.

In what follows, (pre-SNAP) weight updates use equation \ref{eq:generic_weight_update} to update their weights. 
\begin{equation}
\delta W_{ij} = {\alpha} r_{ij}, 
\label{eq:generic_weight_update}
\end{equation}
where $\alpha$ is the learning rate and the learning rule prescribes $r_{ij}$.

\textbf{\subsubsubsection{Hidden Layer Weight Update Rule:}} The hidden layer employs Sanger’s rule \eqref{eq:sanger}\cite{sangersrule}, which extends the basic Hebbian Learning rule introduced in Equation \eqref{eq:hebb}\cite{krotov2016dense}. Basic Hebbian learning updates weights based on the product of the presynaptic activity \(x_j\) and postsynaptic activity \(h_i\), reinforcing the connections between co-active neurons.

\begin{equation}
 r_{ij} = h_i \cdot x_j 
 \label{eq:hebb}
\end{equation}

Sanger's rule builds on this by aiming to make the neurons represent orthogonal features. It sequentially extracts principal components by subtracting the projection onto previously extracted components, as shown in Equation \eqref{eq:sanger}. The term \(\sum_{k=1}^{i-1} h_k w_{kj}\) represents the projection onto the previous outputs, ensuring each neuron’s weight vector remains orthogonal to those of prior neurons.

\begin{equation}
 r_{ij} = h_i x_j - \eta h_i \sum_{k=1}^{i-1} h_k w_{kj} 
 \label{eq:sanger}
\end{equation}

where $\eta$ controls the strength of the orthogonality constraint in Sanger's rule.
\textbf{\subsubsubsection{Classification Layer Weight Update Rule:}} The classification layer in our model employs a supervised Hebbian Learning rule, as defined in Equation \eqref{eq:original_class}. This approach contrasts with the traditional use of Stochastic Gradient Descent (SGD), which is commonly used in other works. Our supervised Hebbian rule offers a novel and more biologically plausible alternative to the standard SGD-based methods, enhancing the neural network’s alignment with biological learning principles.

\begin{equation}
r_{ij} = (\hat{y}_i - \tilde{y}_i) x_j,
\label{eq:original_class}
\end{equation}

\subsection{Derivation of Weight Growth Updates}
\label{app:WeightGrowthDerivation}
Assuming that we have
\begin{align}
    \Wij{T} = \sigma\left(\Wij{0} + \sum_{t=1}^T \dWij{t}\right), 
    \label{app:eq:sigmoid_growth_integral-app}
\end{align}
where $\sigma(x) = \frac{1}{1+e^{-x}}$ is the sigmoid function. Then we have that 
\begin{align}
    &\Wij{T} - \Wij{T-1} & \approx & \sigma^\prime\left(\Wij{0} + \sum_{t=1}^{T-1} \dWij{t}\right)\dWij{T} \label{app:eq:sig-dev-approx}&&\\
    && = & \left(1 - \sigma\left(\Wij{0} + \sum_{t=1}^{T-1} \dWij{t}\right)\right)\sigma\left(\Wij{0} + \sum_{t=1}^{T-1} \dWij{t}\right)\dWij{T} \label{app:eq:sig-dev-sigdif}&&\\
    &&=&\left(1 - \Wij{T-1}\right)\Wij{T-1}\dWij{T}, 
    \label{app:eq:sig-dev-final}&&
\end{align}
where \eqref{app:eq:sig-dev-approx} is the first order expansion in the Taylor series, \eqref{app:eq:sig-dev-sigdif} uses the identity that the derivative of the sigmoid is $\sigma^\prime(x) = (1-\sigma(x))\sigma(x)$, and \eqref{app:eq:sig-dev-final} comes from using \eqref{app:eq:sigmoid_growth_integral-app}. 

Similarly, if instead we wish to have exponential weight growth such that
\begin{align}
    \Wij{T} = \exp\left(\Wij{0} + \sum_{t=1}^T \dWij{t}\right). 
    \label{app:eq:exp_growth_integral-app}
\end{align}
Then we have that 
\begin{align}
    &\Wij{T} - \Wij{T-1} & \approx & \exp^\prime\left(\Wij{0} + \sum_{t=1}^{T-1} \dWij{t}\right)\dWij{T} \label{app:eq:exp-dev-approx}&&\\
    && = & \exp\left(\Wij{0} + \sum_{t=1}^{T-1} \dWij{t}\right)\dWij{T} \label{app:eq:exp-dev-sigdif}&&\\
    &&=&\Wij{T-1}\dWij{T}, 
    \label{app:eq:exp-dev-final}&&
\end{align}
where \eqref{app:eq:exp-dev-approx} is the first order expansion in the Taylor series, \eqref{app:eq:exp-dev-sigdif} uses the identity that the derivative of the exponential is the exponential, and \eqref{app:eq:exp-dev-final} comes from using \eqref{app:eq:exp_growth_integral-app}.

\subsection{SGD: Effect of Weight Growth on Forgetting }
\label{app:SGDCatastrophic Forgetting}

In order to see if having a sigmoidal weight growth could also help mitigate catastrophic forgetting when training with SGD, we repeated all our experiments but this time the $\dWij$ in Table \ref{tab:growth-eqns} were obtained using SGD on cross-entropy loss rather through Hebbian learning. The network architecture stayed the same and the applied weight changes were still the $\DWij$ of Table \ref{tab:growth-eqns}. 

\begin{figure}[htbp]
  \centering
  \begin{subfigure}{0.48\textwidth}
    \centering
    \includegraphics[width=\textwidth]{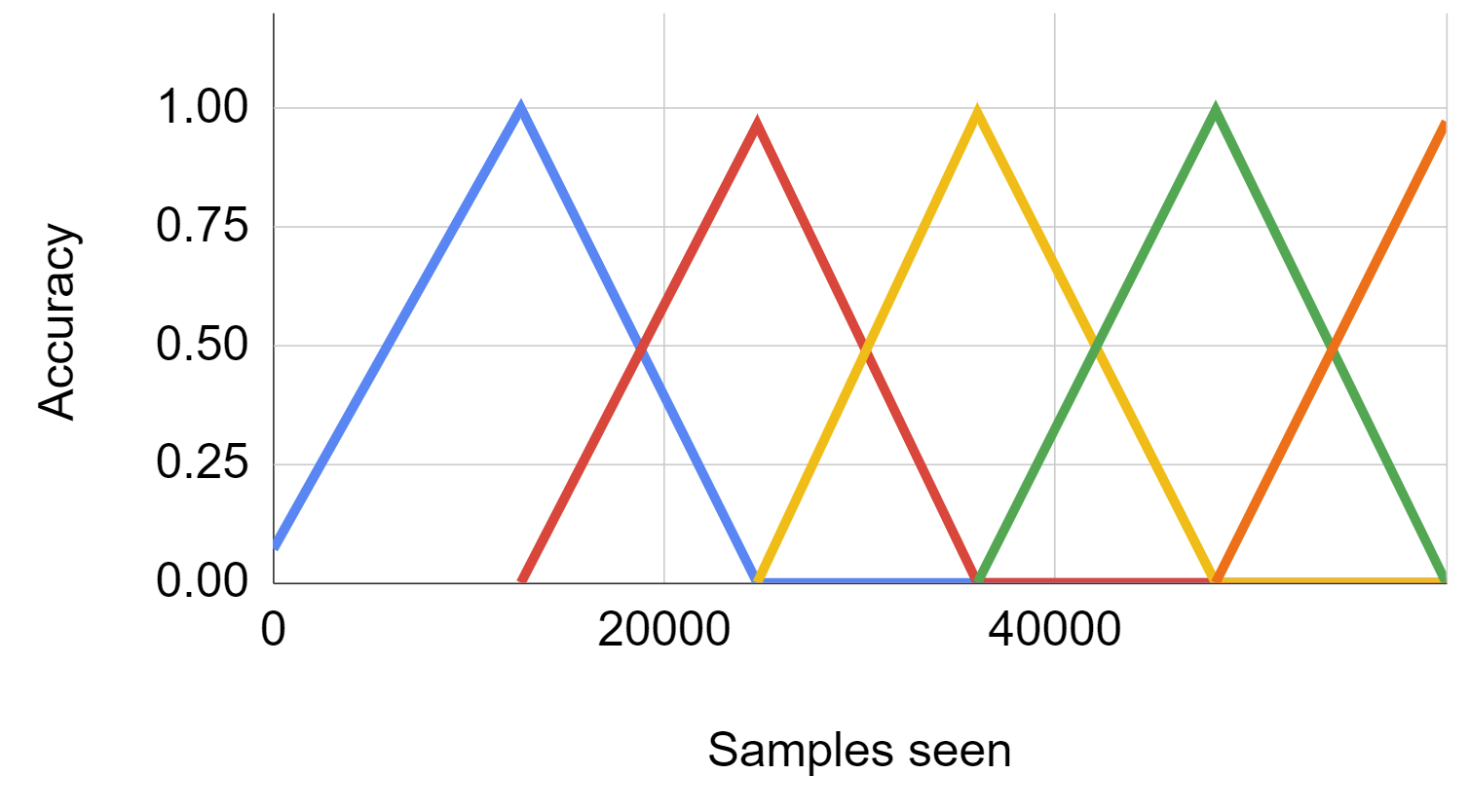}
    \caption{Exponential SGD}
    \label{fig:FORGET-EXP-SGD-MNIST}
  \end{subfigure}
  \hfill
  \begin{subfigure}{0.48\textwidth}
    \centering
    \includegraphics[width=\textwidth]{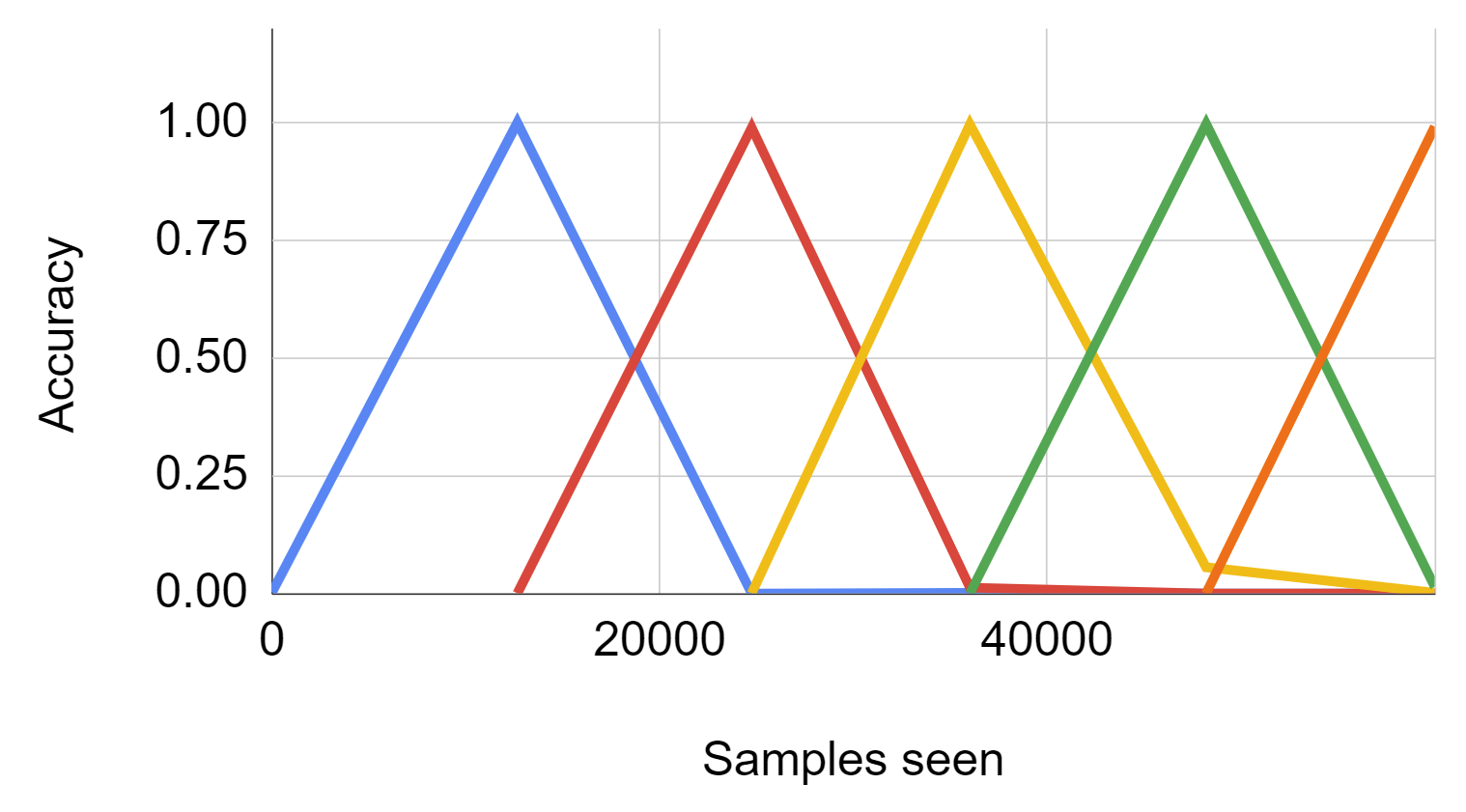}
    \caption{Linear weight growth SGD}
    \label{fig:FORGET-LINEAR-SGD-MNIST}
  \end{subfigure}
  \vfill
  \begin{subfigure}{0.48\textwidth}
    \centering
    \includegraphics[width=\textwidth]{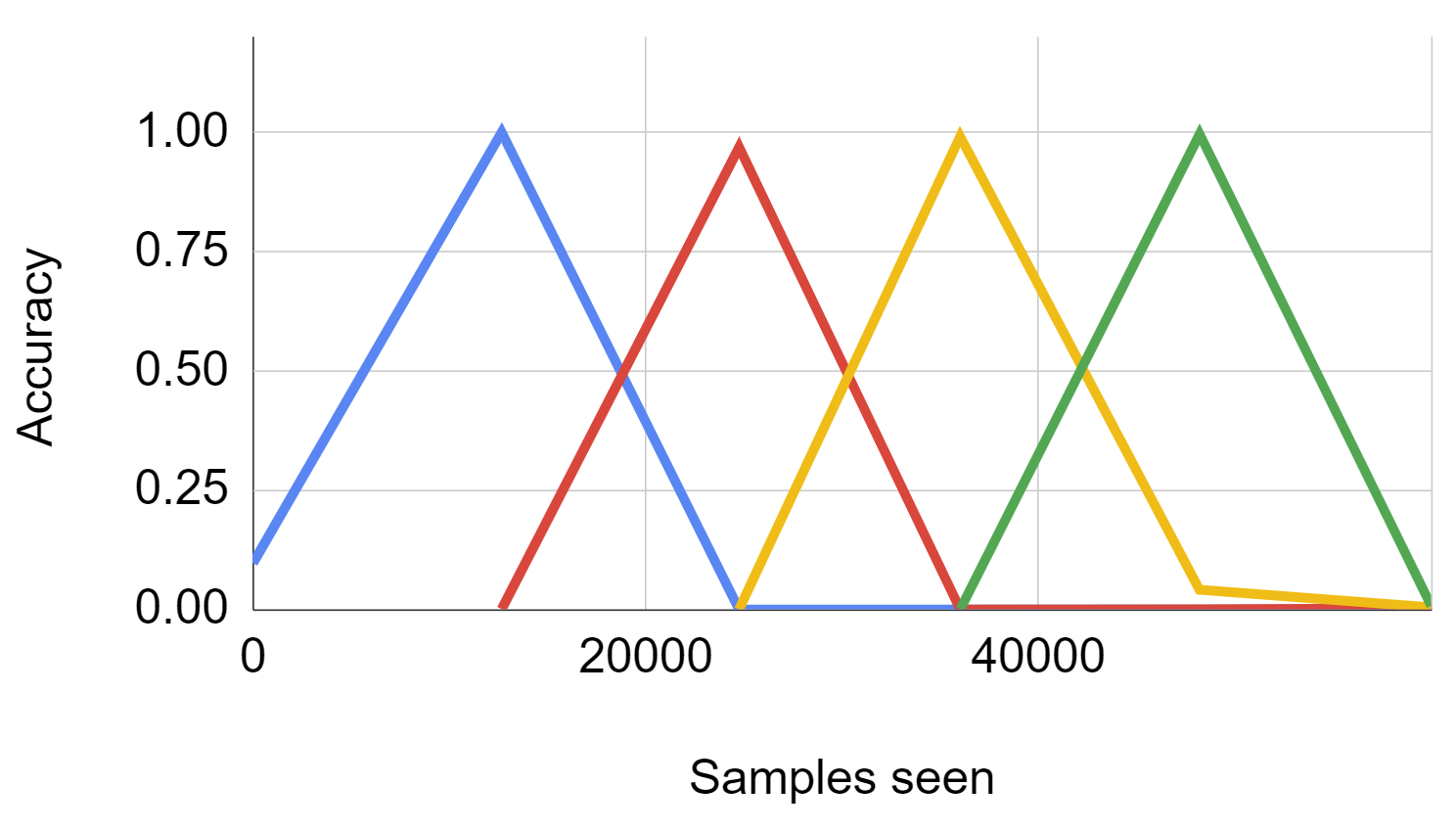}
    \caption{Neuron-wise sigmoidal weight growth SGD}
    \label{fig:FORGET-SIGMOID-NEURON-SGD-MNIST}
  \end{subfigure}
  \hfill
  \begin{subfigure}{0.48\textwidth}
    \centering
    \includegraphics[width=\textwidth]{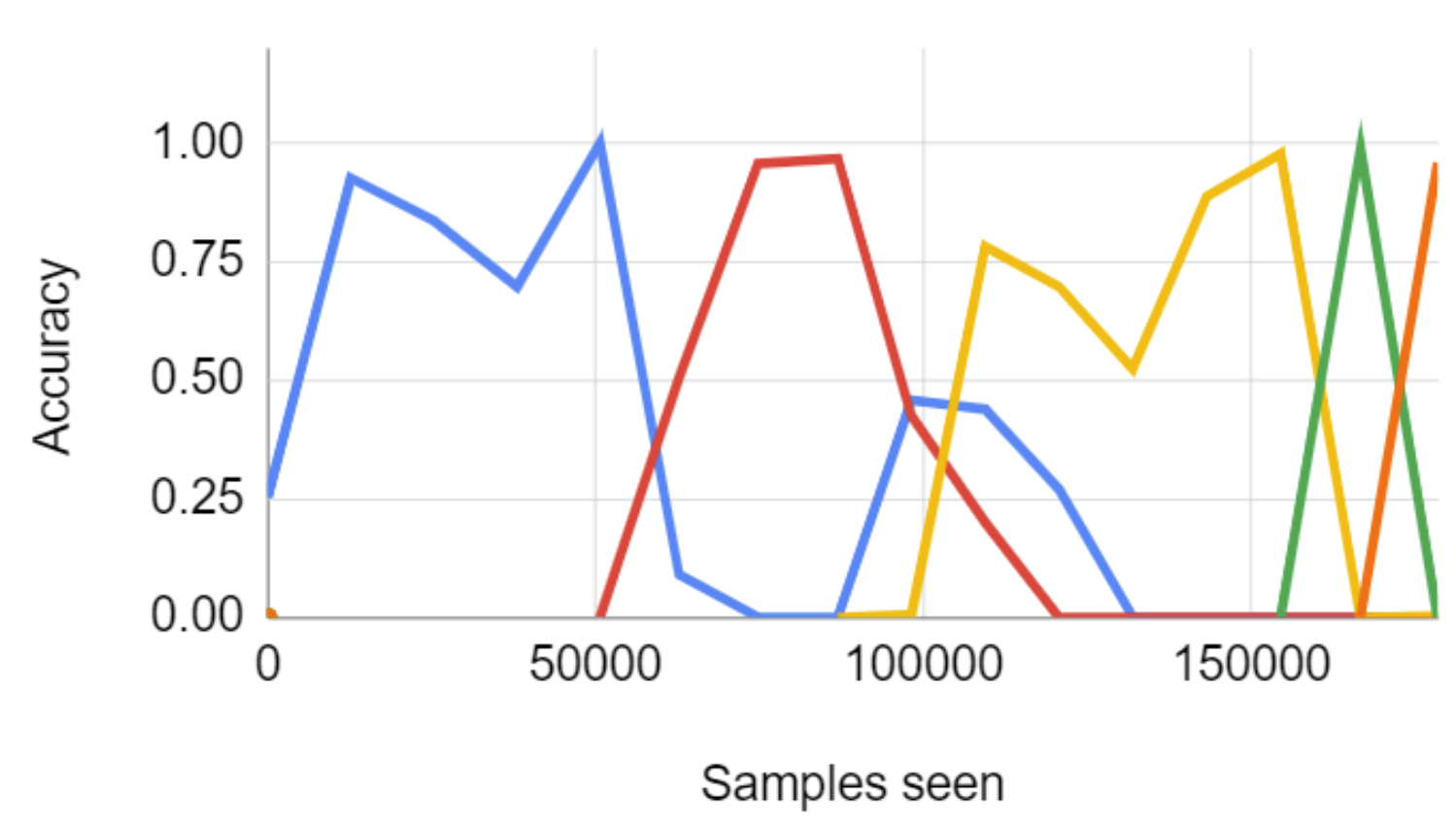}
    \caption{Synapse-wise sigmoidal weight growth SGD}
    \label{fig:FORGET-SIGMOID-SYNAPSE-SGD-MNIST}
  \end{subfigure}
  \begin{subfigure}[b]{1.0\textwidth}
    \centering
    \includegraphics[width=\textwidth]{ExperimentGraphs/MNIST_LEGEND_NUMBER.png}
    \caption{Legend}
    \label{fig:Legend}
  \end{subfigure}
  \caption{Comparison of various \textbf{SGD} models on \textbf{sequential task learning experiments} in \textbf{MNIST} with different hidden and classification layer weight growth functions. \textbf{Legend:} Each line represents the retention of test accuracy across different tasks. \textbf{Tasks:} The blue line indicates the test accuracy on the first task (e.g., \textbf{MNIST} classes 0 and 1), while the orange line shows the test accuracy on the last task}
  \label{fig:mnist_sgd_model_performance}
\end{figure}

\begin{figure}[htbp]
  \centering
  \begin{subfigure}[b]{0.48\textwidth}
    \centering
    \includegraphics[width=\textwidth]{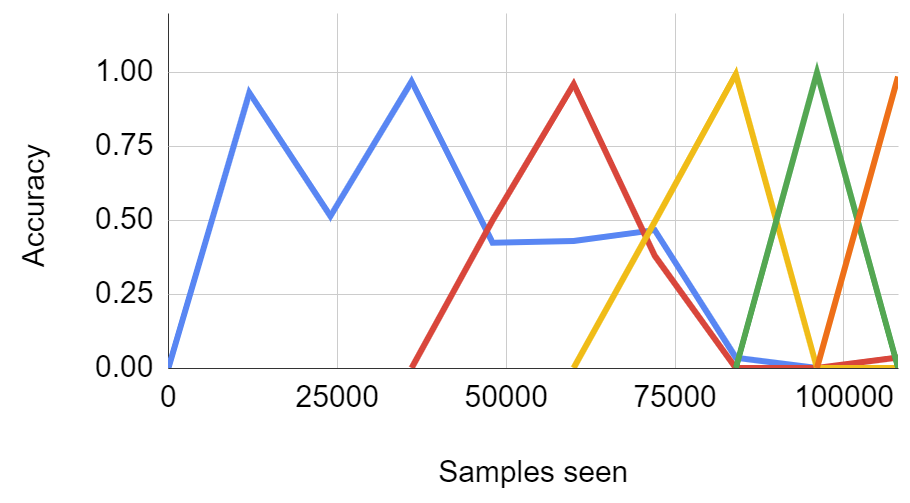}
    \caption{Neuron-wise exponential Hidden and Classification Layer}
    \label{fig:FORGET-EXP-SGD-FMNIST}
  \end{subfigure}
  \hfill
  \begin{subfigure}[b]{0.48\textwidth}
    \centering
    \includegraphics[width=\textwidth]{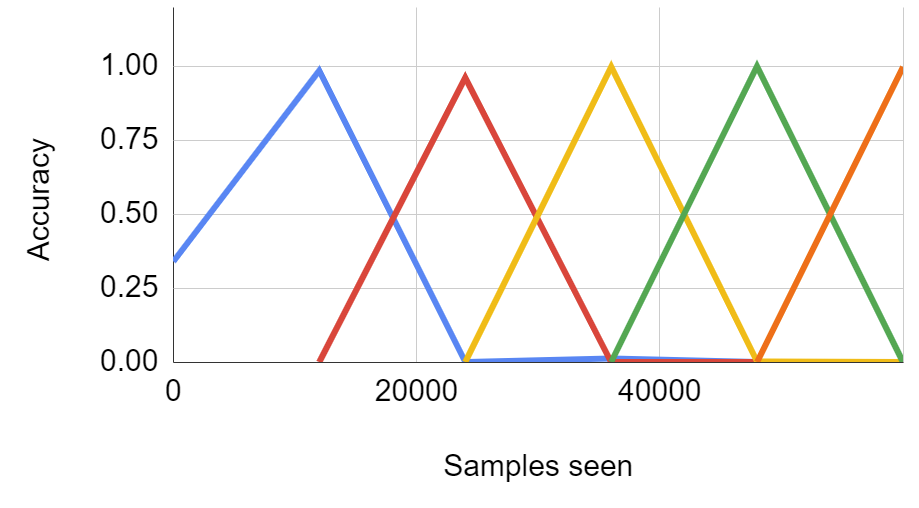}
    \caption{Linear Hidden and Classification Layer}
    \label{fig:FORGET-LINEAR-SGD-FMNIST}
  \end{subfigure}
  \vfill
  \begin{subfigure}[b]{0.48\textwidth}
    \centering
    \includegraphics[width=\textwidth]{ExperimentGraphs/SGD_FORGET/MNIST/SIG_SIG_SGD_NEURON.png}
    \caption{Neuron-wise sigmoid Hidden and Classification Layer}
    \label{fig:FORGET-SIGMOID-NEURON-SGD-FMNIST}
  \end{subfigure}
  \hfill
  \begin{subfigure}[b]{0.48\textwidth}
    \centering
    \includegraphics[width=\textwidth]{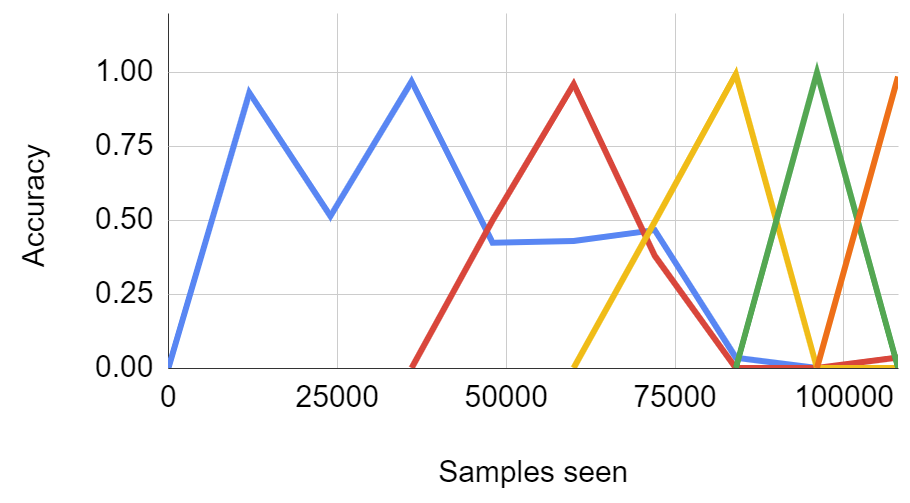}
    \caption{Synapse-wise sigmoid Hidden and Classification Layer}
    \label{fig:FORGET-SIGMOID-SYNAPSE-SGD-FMNIST}
  \end{subfigure}
  \begin{subfigure}[b]{1.0\textwidth}
    \centering
    \includegraphics[width=\textwidth]{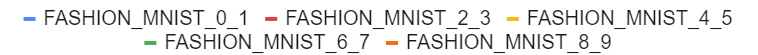}
    \caption{Legend}
    \label{fig:Legend}
  \end{subfigure}
  \caption{Comparison of various \textbf{SGD} models on \textbf{sequential task learning experiments} in \textbf{FashionMNIST} with different hidden and classification layer weight growth functions. \textbf{Legend:} Each line represents the retention of test accuracy across different tasks. \textbf{Tasks:} The blue line indicates the test accuracy on the first task (e.g., \textbf{FashionMNIST} classes 0 and 1), while the orange line shows the test accuracy on the last task}
  \label{fig:Fmnist_sgd_model_performance}
\end{figure}

As can be seen from \fig{fig:mnist_sgd_model_performance} and \fig{fig:Fmnist_sgd_model_performance}, having sigmoidal weight growth does not prevent catastrophic forgetting when learning with SGD, though it slows down the forgetting slightly compared to the usual linear weight growth.

\begin{table}[h!]
\centering
\begin{tabular}{|c|c|c|c|c|}
\hline
\textbf{} & \textbf{Exp Exp neuron} & \textbf{Sig Sig neuron} & \textbf{Sig Sig synapse} & \textbf{Linear Linear} \\ \hline
\textbf{MNIST Accuracy} & 0.9147 & 0.9491 & 0.3531 & 0.9731 \\ \hline
\textbf{FashionMNIST Accuracy} & 0.8333 & 0.8661 & 0.1099 & 0.8717 \\ \hline
\end{tabular}
\caption{Classification accuracies of SGD for different weight growths for MNIST and FashionMNIST dataset in the i.i.d. setting.}
\label{tab:sgd-baseline}
\end{table}

\subsection{Supplementary Graphs}
\label{app:MnistClassDigitRentetion}
\begin{figure}[htbp]
  \centering
  \begin{subfigure}[b]{0.48\textwidth}
    \centering
    \includegraphics[width=\textwidth]{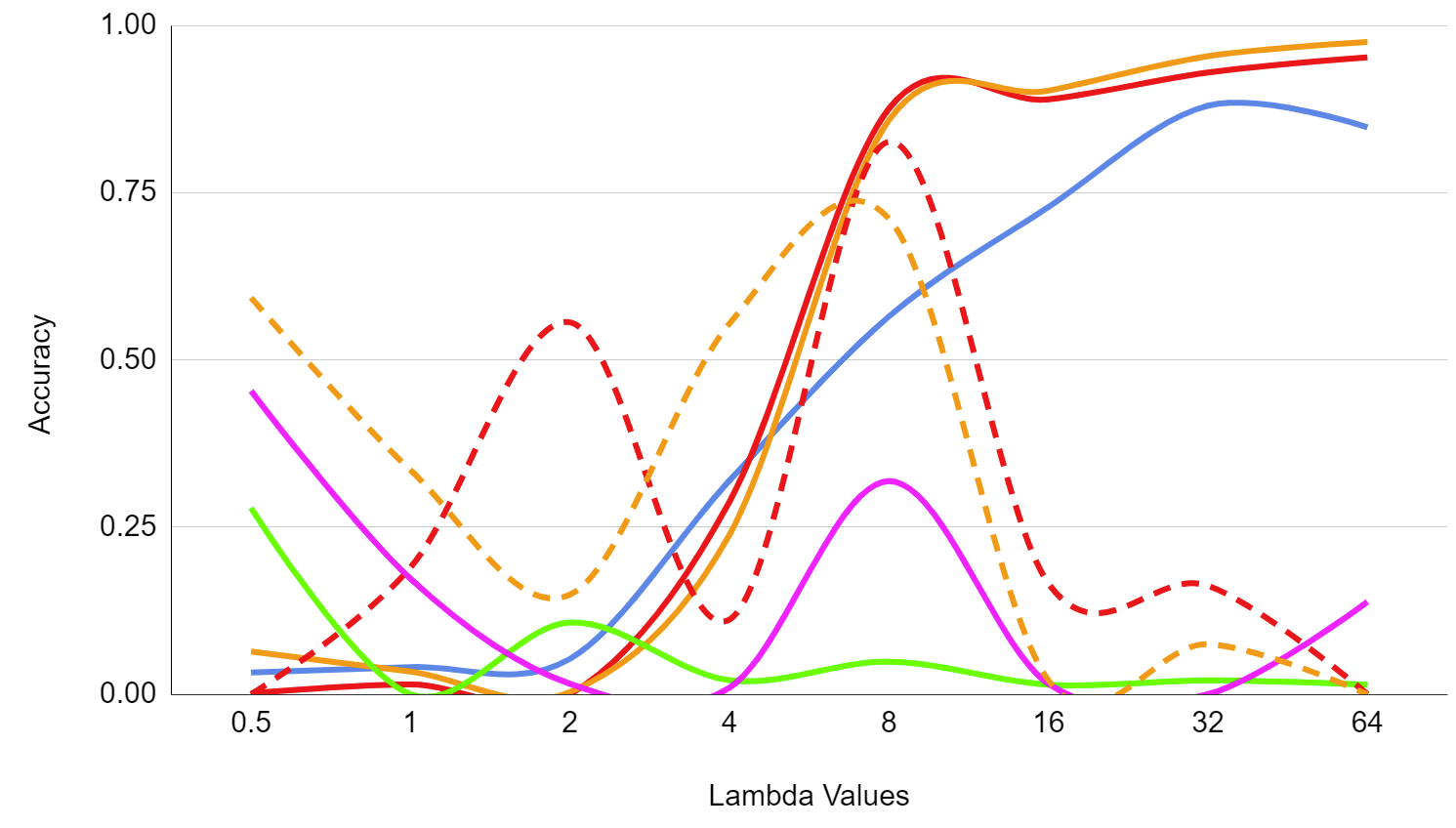}
    \caption{Digits 0 and 1}
    \label{fig:FORGET_OVERALL_0_1}
  \end{subfigure}
  \hfill
  \begin{subfigure}[b]{0.48\textwidth}
    \centering
    \includegraphics[width=\textwidth]{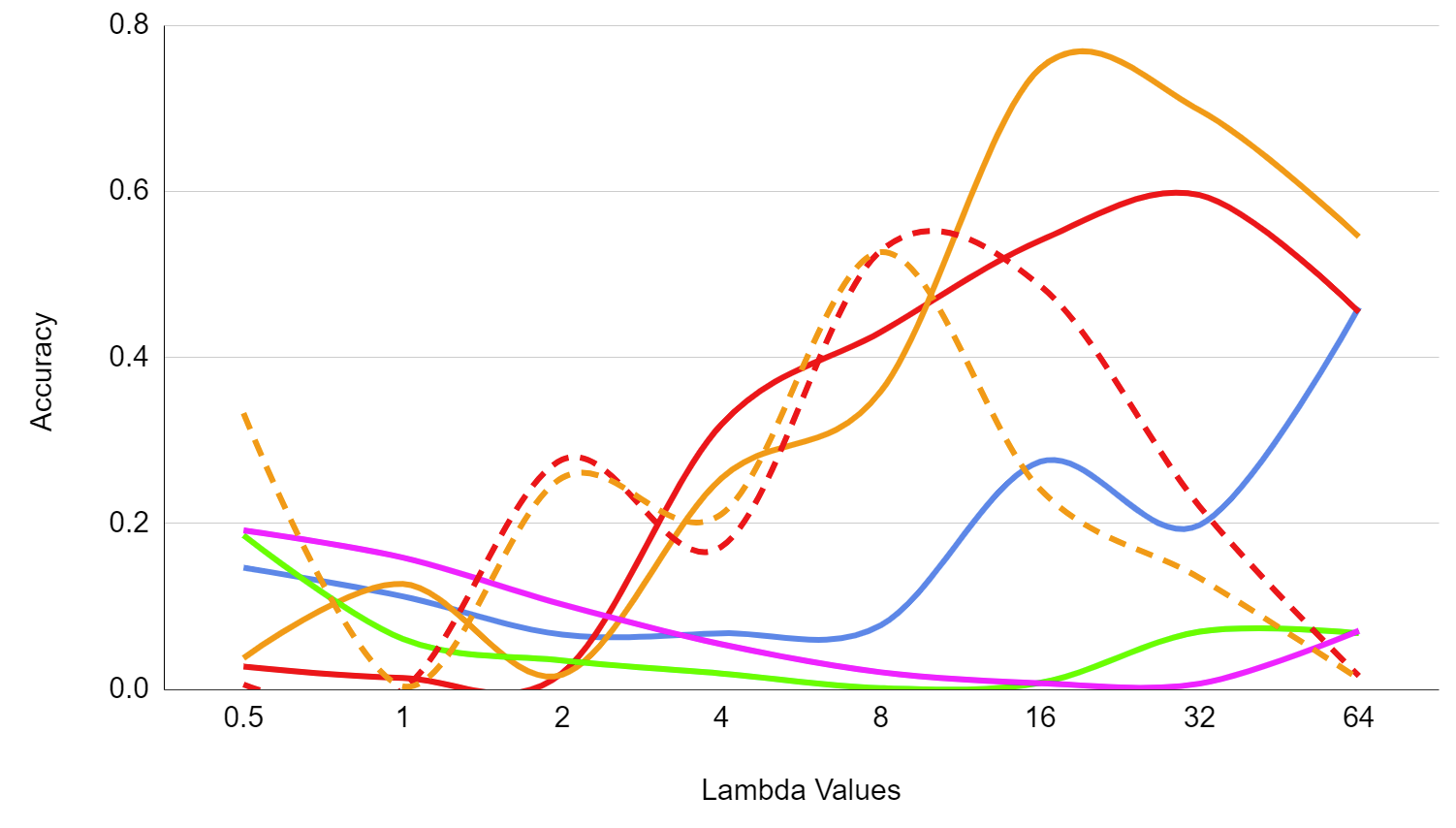}
    \caption{Digits 2 and 3}
    \label{fig:FORGET_OVERALL_2_3}
  \end{subfigure}
  \hfill
  \begin{subfigure}[b]{0.48\textwidth}
    \centering
    \includegraphics[width=\textwidth]{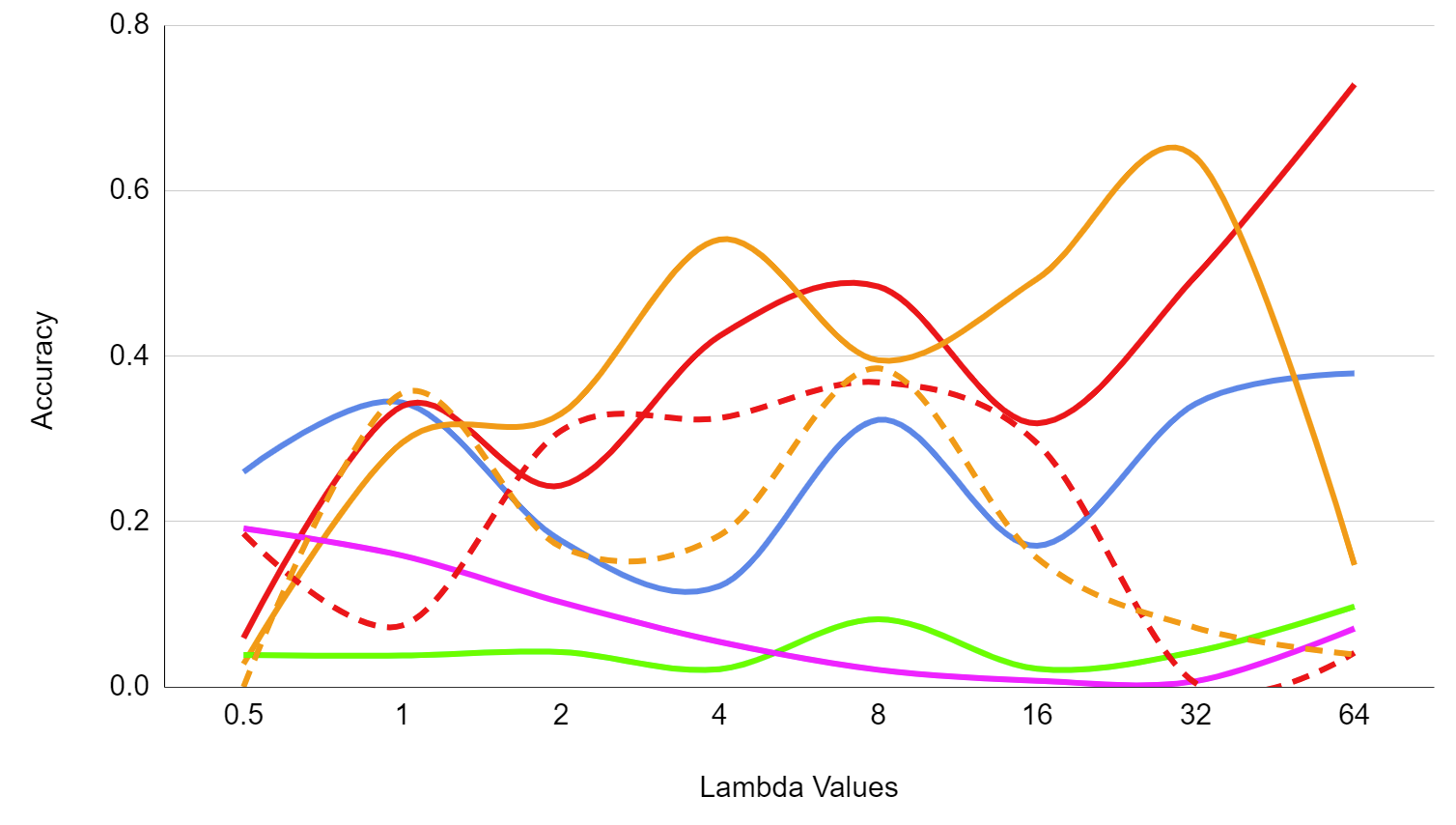}
    \caption{Digits 4 and 5}
    \label{fig:FORGET_OVERALL_4_5}
  \end{subfigure}
  \hfill
  \begin{subfigure}[b]{0.48\textwidth}
    \centering
    \includegraphics[width=\textwidth]{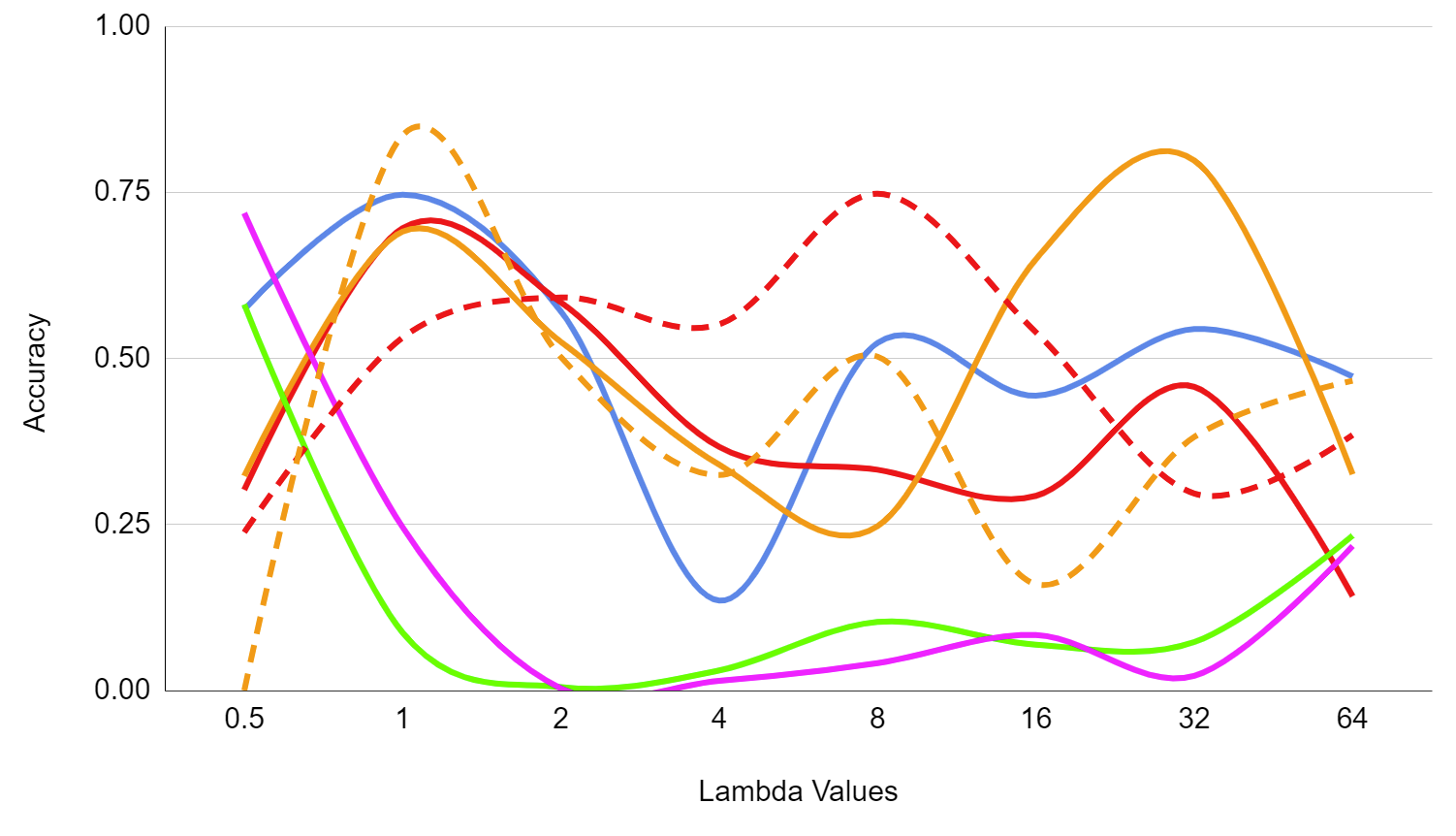}
    \caption{Digits 6 and 7}
    \label{fig:FORGET_OVERALL_6_7}
  \end{subfigure}
  \hfill
  \begin{subfigure}[b]{0.48\textwidth}
    \centering
    \includegraphics[width=\textwidth]{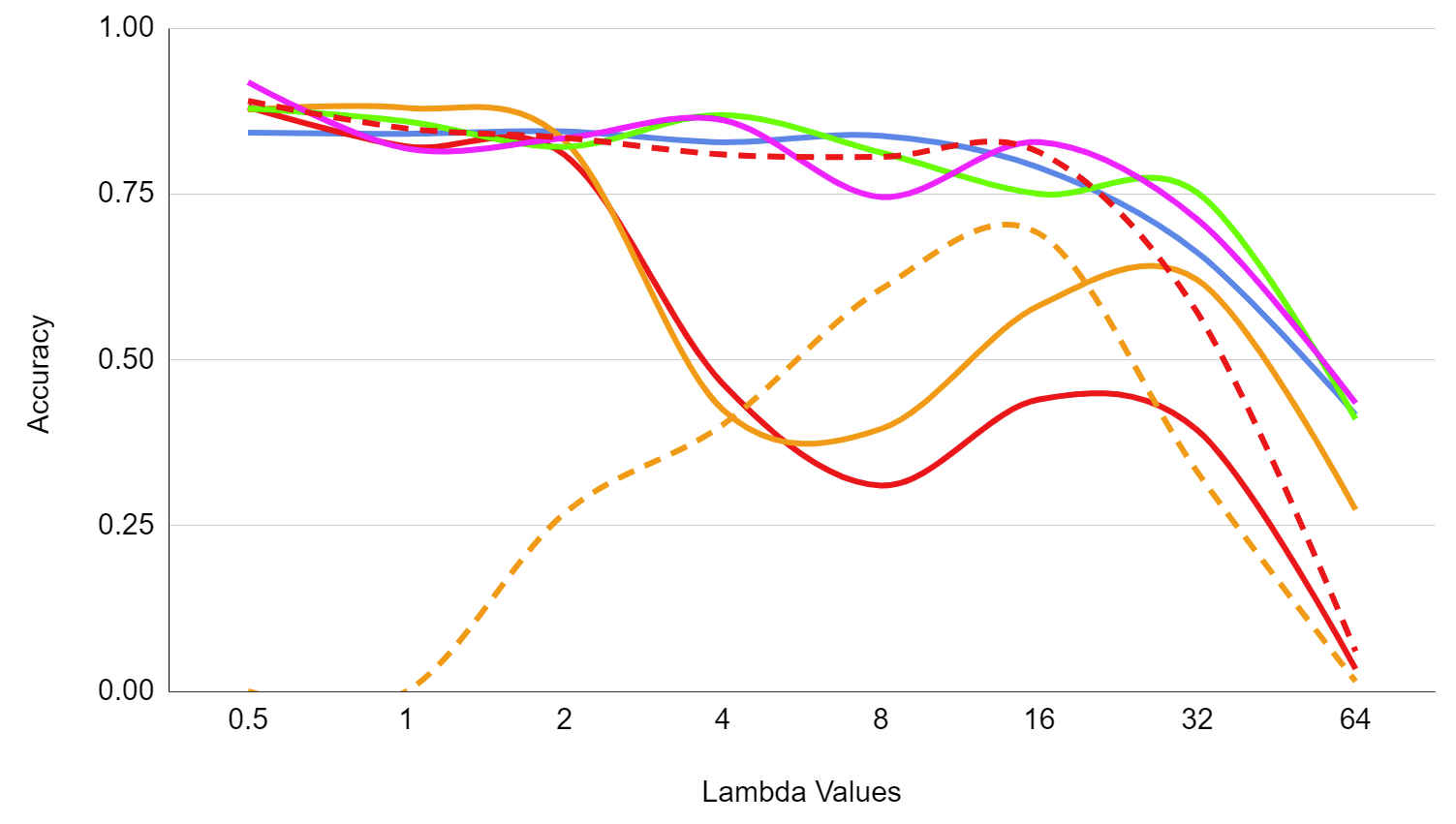}
    \caption{Digits 8 and 9}
    \label{fig:FORGET_OVERALL_8_9}
  \end{subfigure}
  \vspace{1em} 
  
  \begin{subfigure}[b]{1.0\textwidth}
    \centering
    \includegraphics[width=\textwidth]{ExperimentGraphs/legend_function_type.png}
    \caption{Legend}
    \label{fig:Legend}
  \end{subfigure}
  \caption{Retention accuracy of \textbf{Hebbian} MLP on the \textbf{sequential task} versions for different digit pairs in \textbf{MNIST} as a function of lateral inhibition hyperparameter $\lambda$ and the type of weight growth. \textbf{Legend:} The first term specifies the weight growth type of the hidden layer, and the second specifies the weight growth type of the output layer. For example, red lines denote sigmoidal growth for the hidden layer and linear growth for the output layer. Solid lines indicate neuron-wise growth, while dotted lines indicate synapse-wise growth (cf. Table \ref{tab:growth-eqns}).}
  \label{fig:DIGIT_Forget_MNIST_Model_Performance}
\end{figure}

\label{app:MnistTopPerformers}
\begin{figure}[htbp]
  \centering
  \begin{subfigure}[b]{0.48\textwidth}
    \centering
    \includegraphics[width=\textwidth]{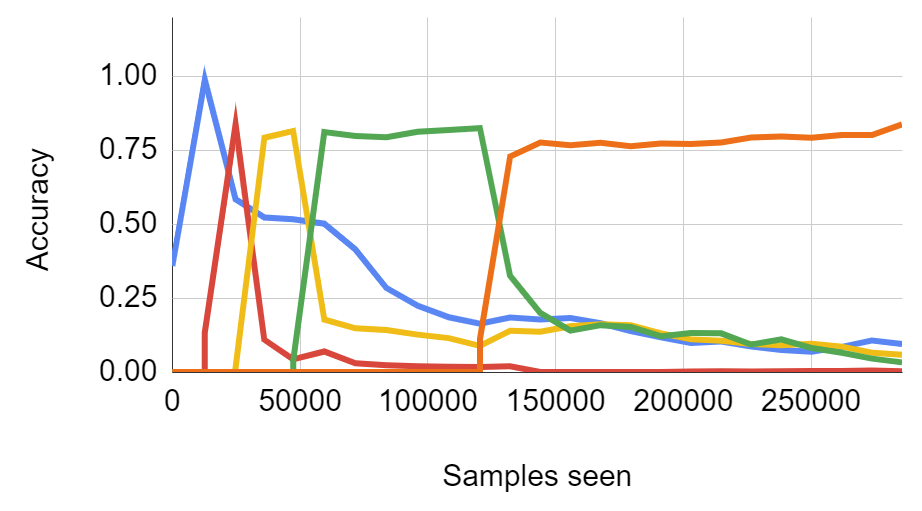}
    \caption{Exponential Hidden and Classification Layer, Neuron-wise, $\lambda = 8$}
    \label{fig:mnist_hebb_FORGET-EXP-EXP-NEURON-8}
  \end{subfigure}
  \hfill
  \begin{subfigure}[b]{0.48\textwidth}
    \centering
    \includegraphics[width=\textwidth]{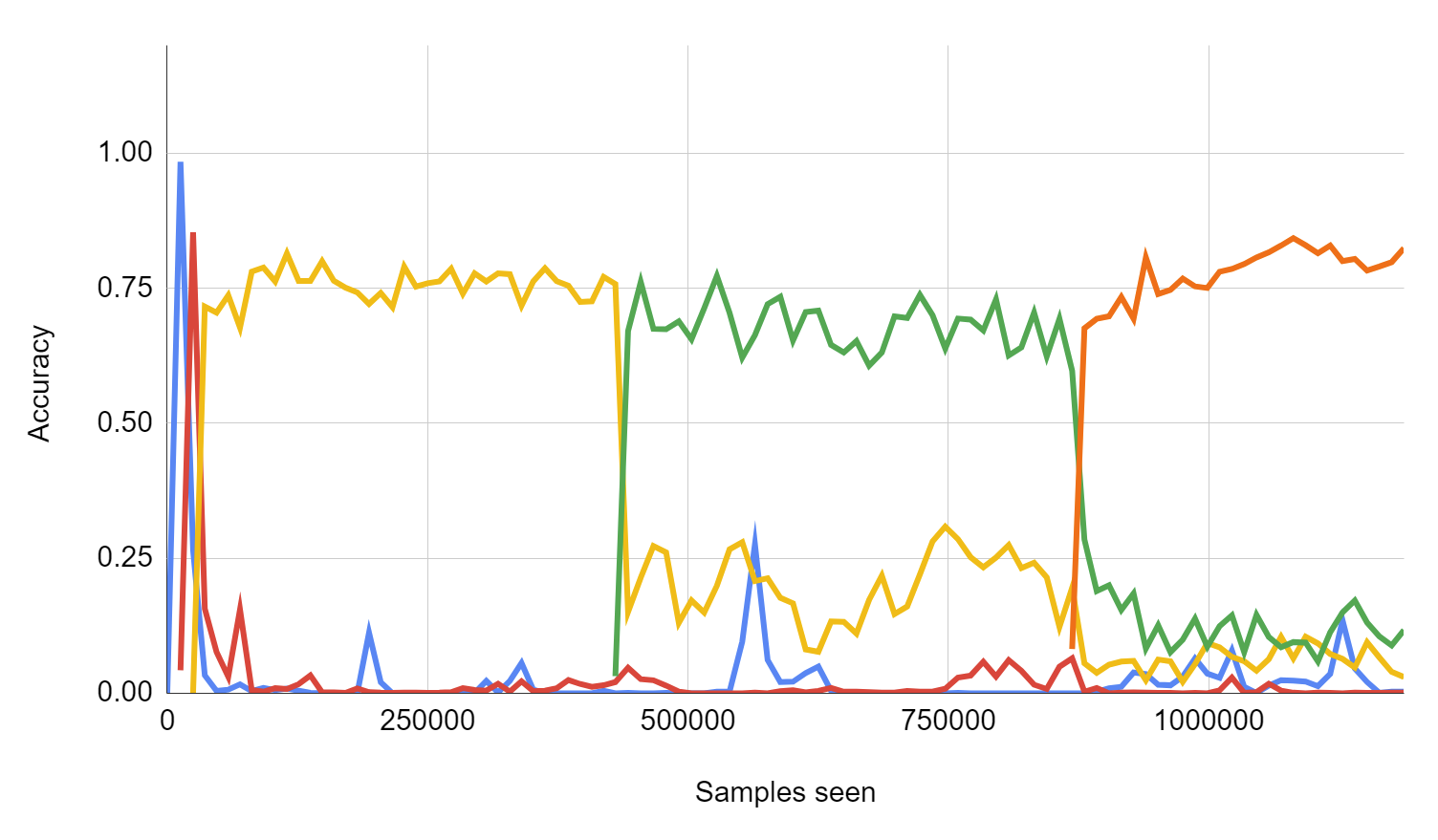}
    \caption{Exponential Hidden, Linear Classification Layer, Neuron-wise, $\lambda = 16$}
    \label{fig:mnist_hebb_FORGET_EXP_LINEAR_16}
  \end{subfigure}
  \vfill
  \begin{subfigure}[b]{0.48\textwidth}
    \centering
    \includegraphics[width=\textwidth]{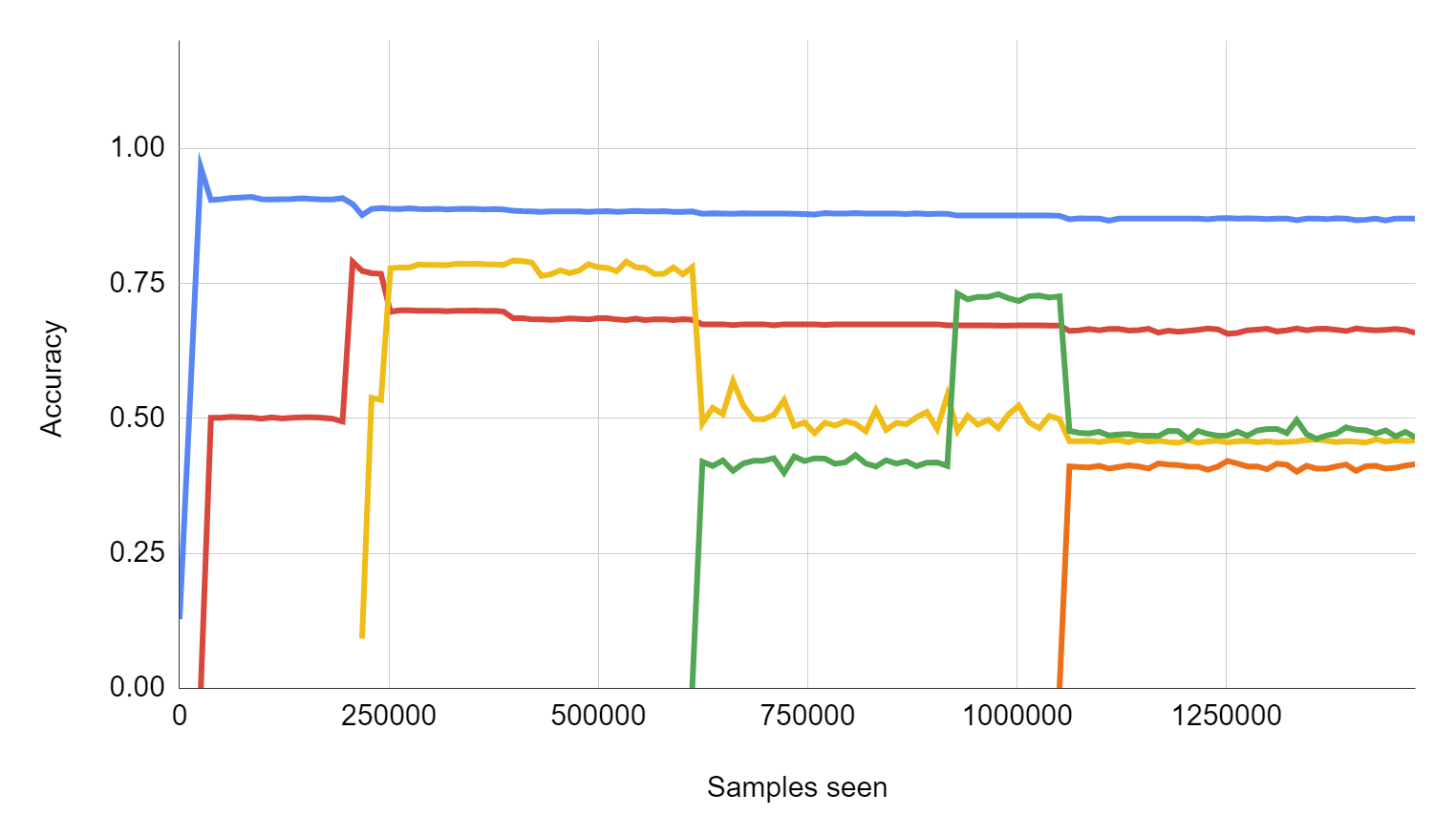}
    \caption{Sigmoid Hidden, Linear Classification Layer, Neuron-wise, $\lambda = 32$}
    \label{fig:mnist_hebb_FORGET_SIG_LINEAR_NEURON_32}
  \end{subfigure}
  \hfill
  \begin{subfigure}[b]{0.48\textwidth}
    \centering
    \includegraphics[width=\textwidth]{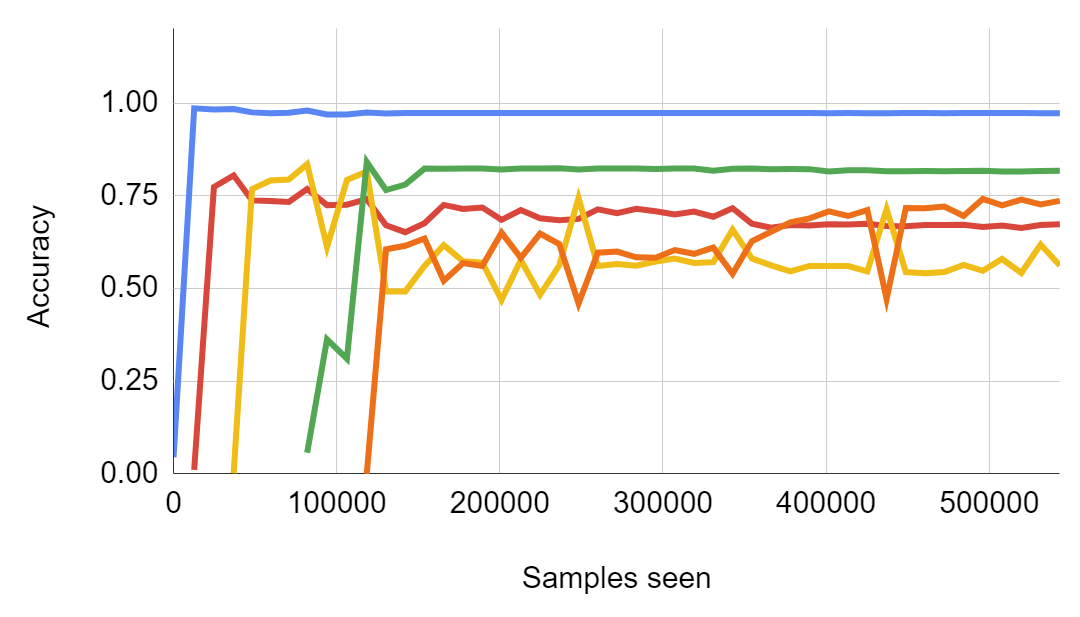}
    \caption{Sigmoid Hidden and Classification Layer, Neuron-wise, $\lambda = 32$}
    \label{fig:mnist_hebb_FORGET_SIG_SIG_NEURON_32}
  \end{subfigure}
  \vfill
  \begin{subfigure}[b]{0.48\textwidth}
    \centering
    \includegraphics[width=\textwidth]{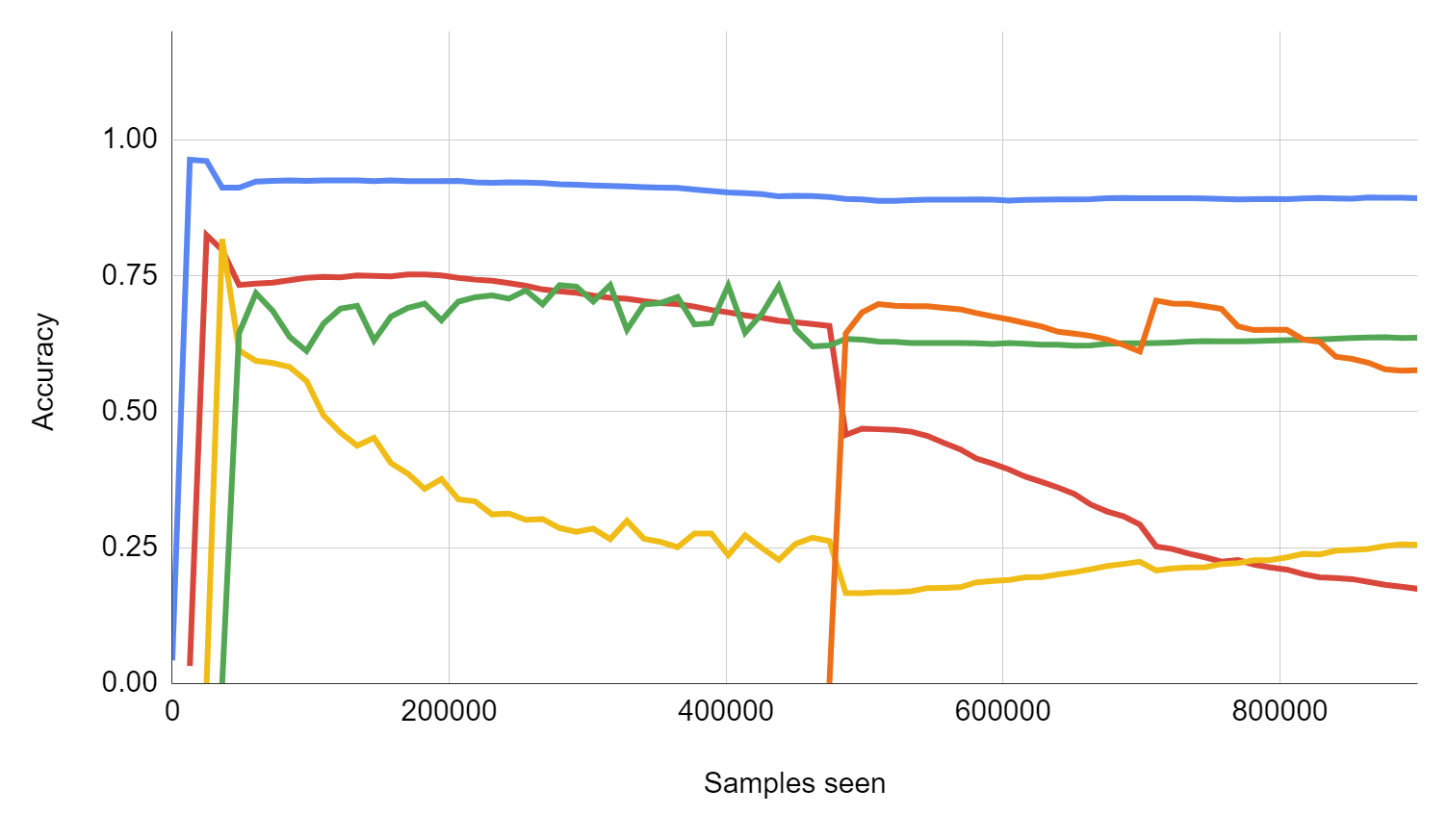}
    \caption{Linear Hidden and Classification Layer, $\lambda = 32$}
    \label{fig:mnist_hebb_FORGET_LINEAR_LINEAR_NEURON_32}
  \end{subfigure}
  \vspace{1em} 
  
  \begin{subfigure}[b]{1.0\textwidth}
    \centering
    \includegraphics[width=\textwidth]{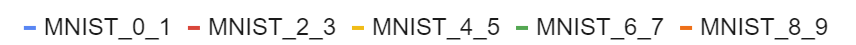}
    \caption{Legend}
    \label{fig:Legend}
  \end{subfigure}
  \caption{Comparison of top-performing \textbf{Hebbian} models on \textbf{sequential task learning experiments} with various hidden and classification layer weight growth functions in \textbf{MNIST}. \textbf{Legend:} Each line represents the retention of test accuracy across different tasks. \textbf{Tasks:} The blue line indicates the test accuracy on the first task (e.g., \textbf{MNIST} classes 0 and 1), while the orange line shows the test accuracy on the last task}
  \label{fig:mnist-hebb_top_performance}
\end{figure}

\label{app:FMnistClassDigitRentetion}
\begin{figure}[htbp]
  \centering
  \begin{subfigure}[b]{0.48\textwidth}
    \centering
    \includegraphics[width=\textwidth]{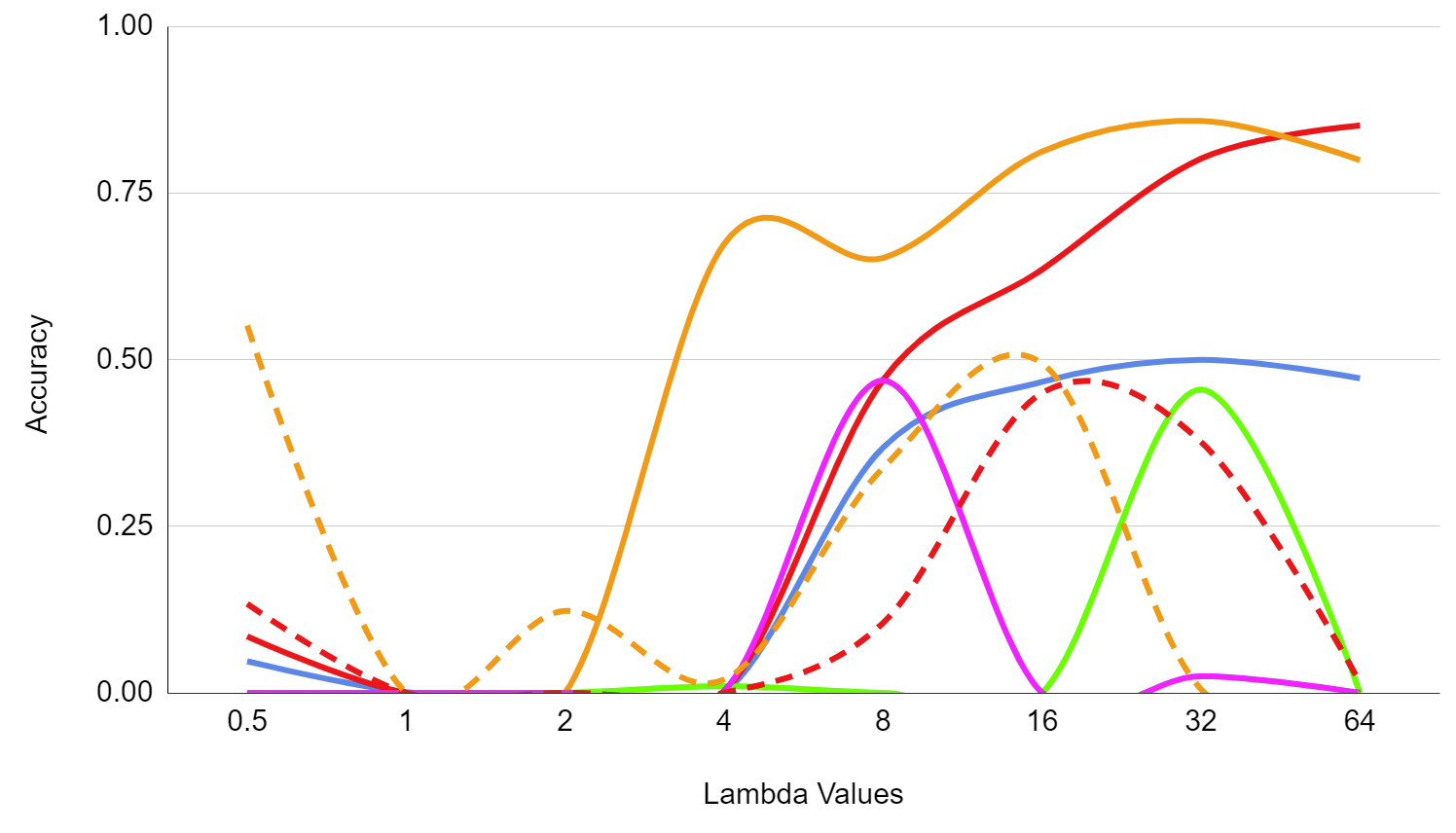}
    \caption{FashionMNIST Classes 0 and 1}
    \label{fig:FMNIST_FORGET_OVERALL_0_1}
  \end{subfigure}
  \hfill
  \begin{subfigure}[b]{0.48\textwidth}
    \centering
    \includegraphics[width=\textwidth]{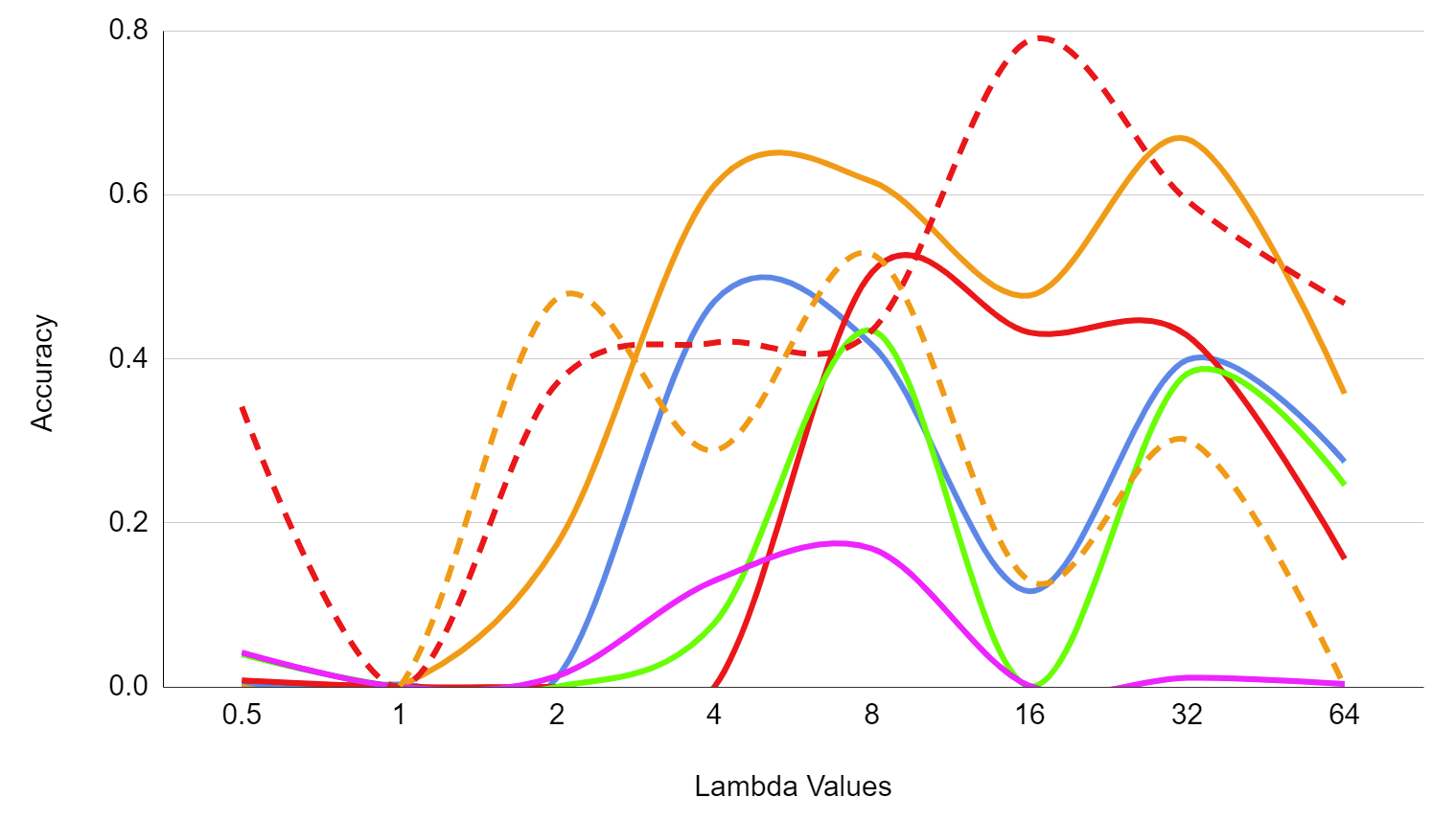}
    \caption{FashionMNIST Classes 2 and 3}
    \label{fig:FMNIST_FORGET_OVERALL_2_3}
  \end{subfigure}
  \hfill
  \begin{subfigure}[b]{0.48\textwidth}
    \centering
    \includegraphics[width=\textwidth]{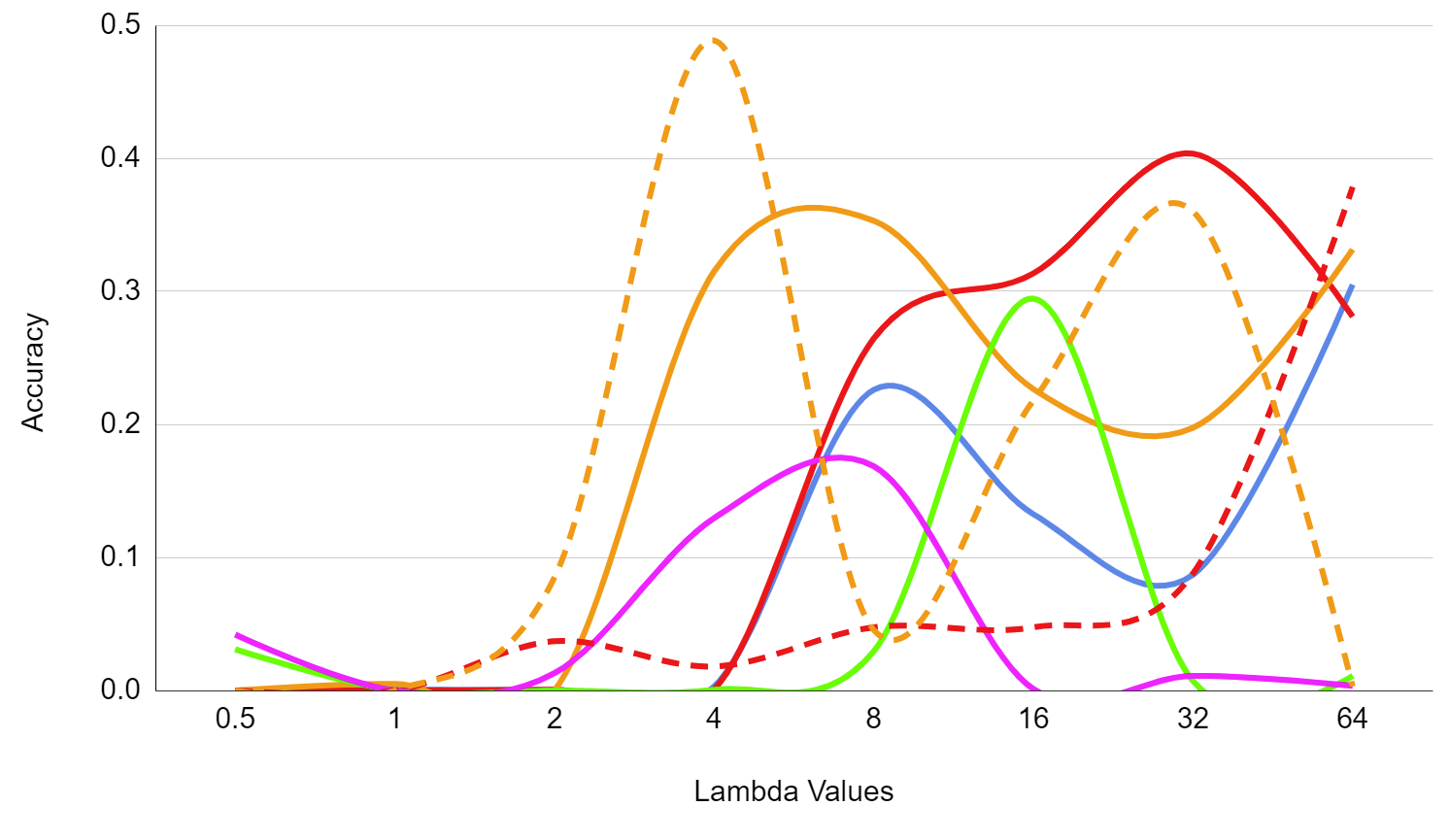}
    \caption{FashionMNIST Classes 4 and 5}
    \label{fig:FMNIST_FORGET_OVERALL_4_5}
  \end{subfigure}
  \hfill
  \begin{subfigure}[b]{0.48\textwidth}
    \centering
    \includegraphics[width=\textwidth]{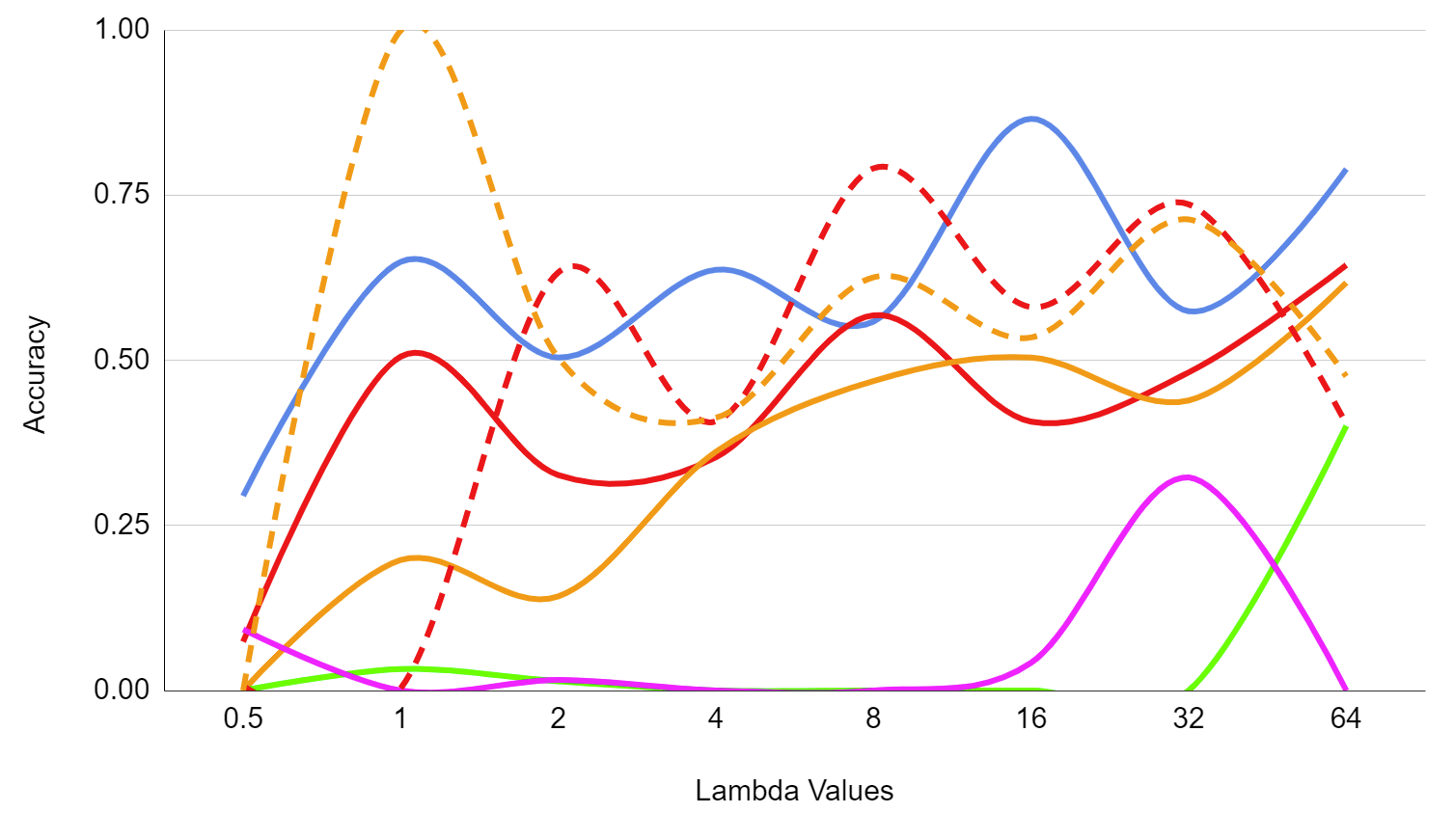}
    \caption{FashionMNIST Classes 6 and 7}
    \label{fig:FMNIST_FORGET_OVERALL_6_7}
  \end{subfigure}
  \hfill
  \begin{subfigure}[b]{0.48\textwidth}
    \centering
    \includegraphics[width=\textwidth]{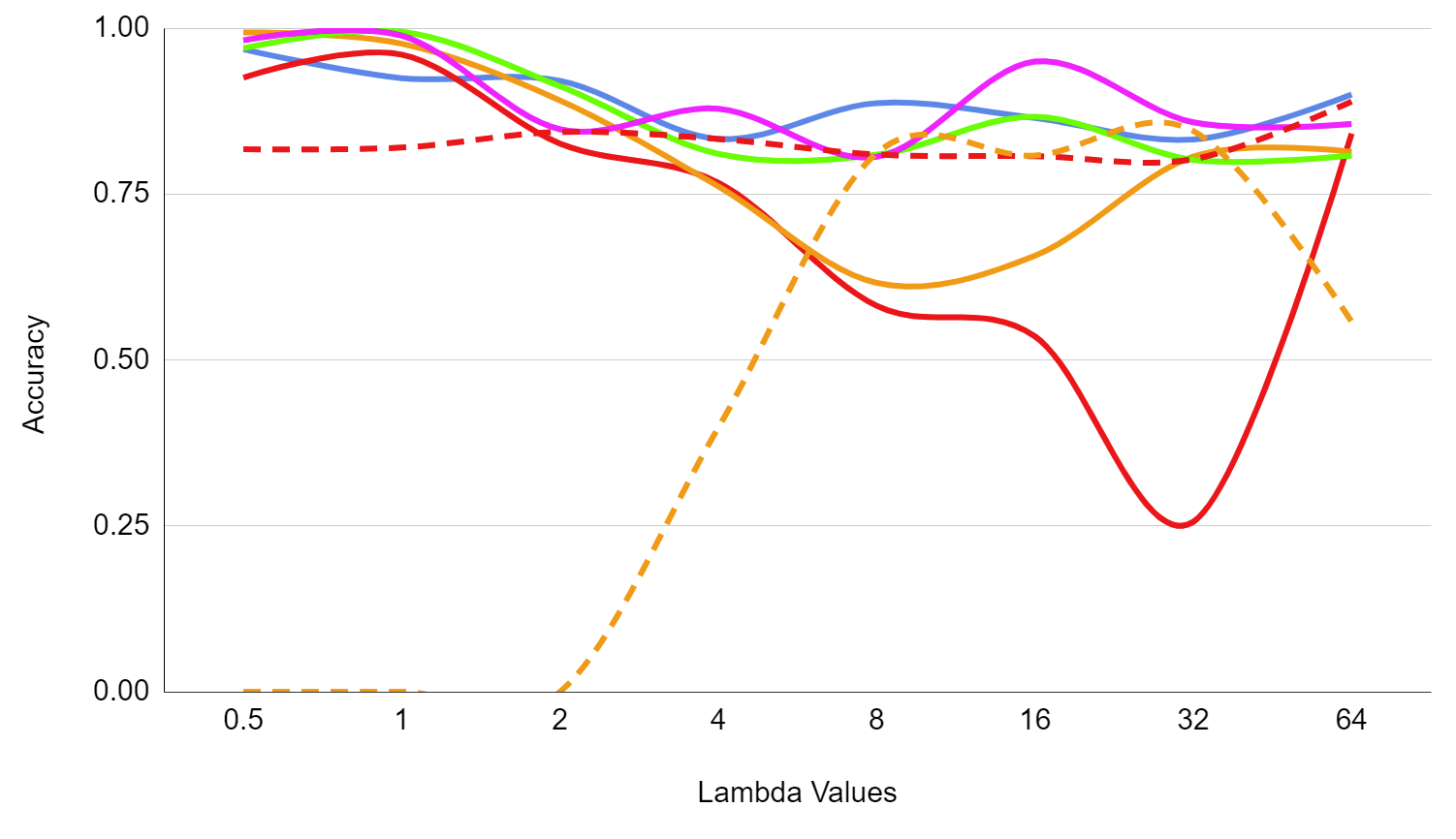}
    \caption{FashionMNIST Classes 8 and 9}
    \label{fig:FMNIST_FORGET_OVERALL_8_9}
  \end{subfigure}
  \vspace{1em} 
  
  \begin{subfigure}[b]{1.0\textwidth}
    \centering
    \includegraphics[width=\textwidth]{ExperimentGraphs/legend_function_type.png}
    \caption{Legend}
    \label{fig:Legend}
  \end{subfigure}
  
  \caption{Retention accuracy of \textbf{Hebbian} MLP on the \textbf{sequential task} versions for different class pairs in \textbf{FashionMNIST} as a function of lateral inhibition hyperparameter $\lambda$ and the type of weight growth. \textbf{Legend:} The first term specifies the weight growth type of the hidden layer, and the second specifies the weight growth type of the output layer. Solid lines represent neuron-wise weight growth, and dotted lines represent synapse-wise weight growth, reflecting how different configurations impact retention across class pairs (cf. Table \ref{tab:growth-eqns}).}
  \label{fig:FMNIST_Forget_Model_Performance}
\end{figure}

\label{app:FMnistTopPerformers}
\begin{figure}[htbp]
  \centering
  \begin{subfigure}{0.48\textwidth}
    \centering
    \includegraphics[width=\textwidth]{ExperimentGraphs/FORGET_EXP/FORGET-EXP-EXP-NEURON-8.png}
    \caption{Exponential Hidden and Classification Layer, Neuron-wise, Lambda 8}
    \label{fig:fmnist_hebb_FORGET-EXP-EXP-NEURON-8}
  \end{subfigure}
  \hfill
  \begin{subfigure}{0.48\textwidth}
    \centering
    \includegraphics[width=\textwidth]{ExperimentGraphs/FORGET_EXP/FORGET_EXP_LINEAR_16.png}
    \caption{Exponential Hidden, Linear Classification Layer, Neuron-wise, Lambda 16}
    \label{fig:fmnist_hebb_FORGET_EXP_LINEAR_16}
  \end{subfigure}
  \vfill
  \begin{subfigure}{0.48\textwidth}
    \centering
    \includegraphics[width=\textwidth]{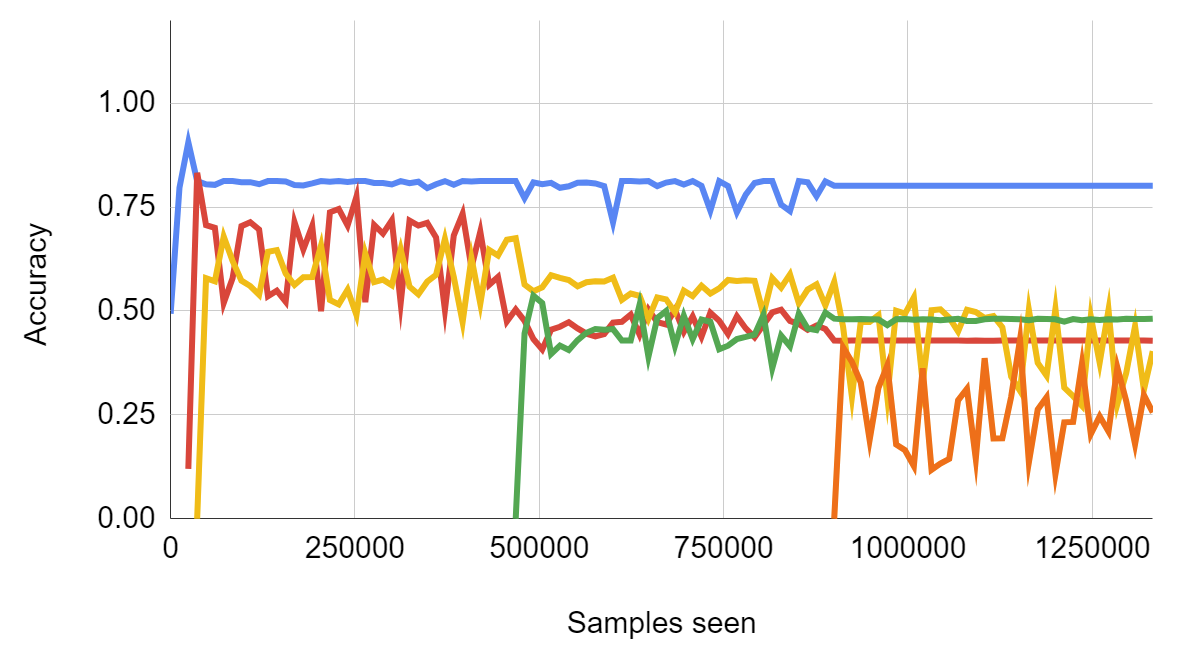}
    \caption{Sigmoid Hidden, Linear Classification Layer, Neuron-wise,  Lambda 32}
    \label{fig:fmnist_hebb_FORGET_SIG_LINEAR_NEURON_32}
  \end{subfigure}
  \hfill
  \begin{subfigure}{0.48\textwidth}
    \centering
    \includegraphics[width=\textwidth]{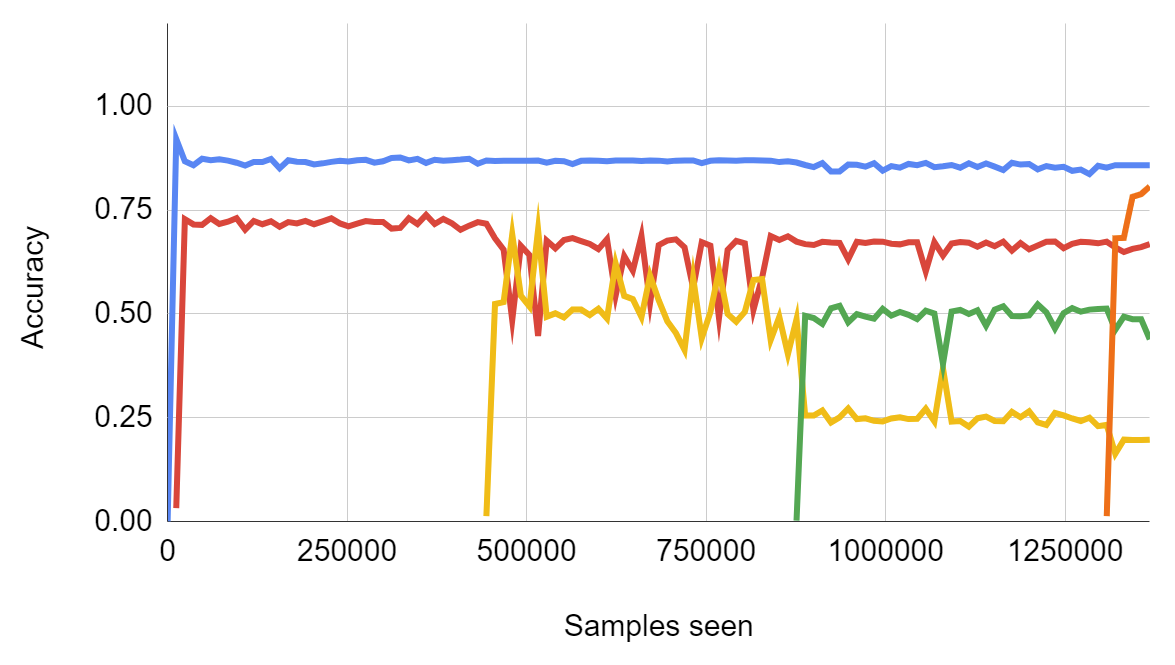}
    \caption{Sigmoid Hidden and Classification Layer, Neuron-wise, Lambda 32}
    \label{fig:fmnist_hebb_FORGET_SIG_SIG_NEURON_32}
  \end{subfigure}
  \vfill
  \begin{subfigure}{0.48\textwidth}
    \centering
    \includegraphics[width=\textwidth]{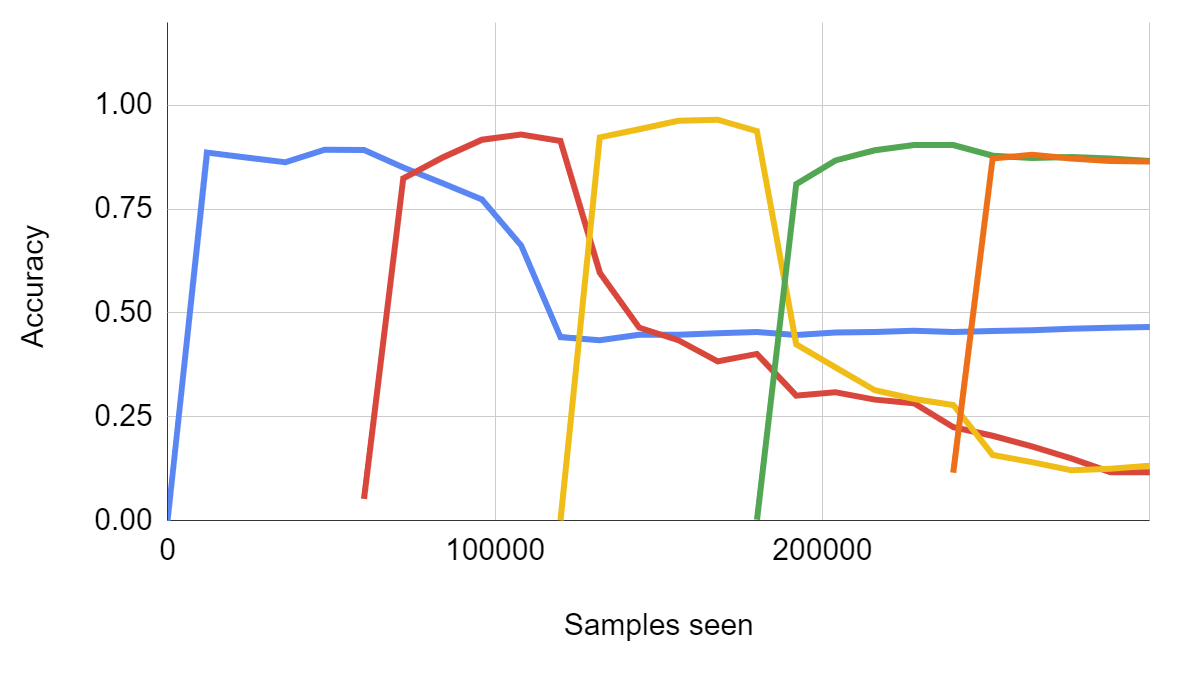}
    \caption{Linear Hidden and Classification Layer, Lambda 16}
    \label{fig:fmnist_hebb_FORGET_LINEAR_LINEAR_NEURON_16}
  \end{subfigure}
  \begin{subfigure}[b]{1.0\textwidth}
    \centering
    \includegraphics[width=\textwidth]{ExperimentGraphs/FMNIST_LEGEND_NUMBER.png}
    \caption{Legend}
    \label{fig:Legend}
  \end{subfigure}
    \caption{Comparison of top-performing \textbf{Hebbian} models on \textbf{sequential task learning experiments} with various hidden and classification layer weight growth functions in \textbf{FashionMNIST}. \textbf{Legend:} Each line represents the retention of test accuracy across different tasks. \textbf{Tasks:} The blue line indicates the test accuracy on the first task (e.g., \textbf{FashionMNIST} classes 0 and 1), while the orange line shows the test accuracy on the last task}
  \label{fig:fmnist_hebb_top_performance}
\end{figure}

\end{document}